\documentclass[10pt,twocolumn,letterpaper]{article}

\usepackage[pagenumbers]{cvpr} %

\usepackage[accsupp]{axessibility}

\usepackage{graphicx}
\usepackage{amsmath}
\usepackage{amssymb}
\usepackage{booktabs}
\usepackage{cite}

\usepackage{algorithm}
\usepackage{algorithmic}
\usepackage{mathtools}

\newlength\savewidth\newcommand\shline{\noalign{\global\savewidth\arrayrulewidth
  \global\arrayrulewidth 1pt}\hline\noalign{\global\arrayrulewidth\savewidth}}

\usepackage{color}
\usepackage{xcolor}
\definecolor{citecolor}{HTML}{0071bc}
\usepackage[pagebackref=true,breaklinks=true,colorlinks,citecolor=citecolor,urlcolor=citecolor,bookmarks=false]{hyperref}

\usepackage[capitalize]{cleveref}
\crefname{section}{Sec.}{Secs.}
\Crefname{section}{Section}{Sections}
\Crefname{table}{Table}{Tables}
\crefname{table}{Tab.}{Tabs.}

\def\confName{CVPR}
\def\confYear{2022}

\begin{document}

\title{GIFS: Neural Implicit Function for General Shape Representation}

\author{Jianglong Ye$^{1}$ \qquad Yuntao Chen$^{2}$\qquad Naiyan Wang$^{2}$\qquad Xiaolong Wang$^{1}$ \\
  $^{1}$UC San Diego\qquad $^{2}$TuSimple }

\maketitle

\begin{abstract}
  Recent development of neural implicit function has shown tremendous success on high-quality 3D shape reconstruction.
  However, most works divide the space into inside and outside of the shape, which limits their representing power to single-layer and watertight shapes.
  This limitation leads to tedious data processing (converting non-watertight raw data to watertight) as well as the incapability of representing general object shapes in the real world.
  In this work, we propose a novel method to represent general shapes including non-watertight shapes and shapes with multi-layer surfaces.
  We introduce General Implicit Function for 3D Shape (GIFS), which models the relationships between every two points instead of the relationships between points and surfaces. Instead of dividing 3D space into predefined inside-outside regions, GIFS encodes whether two points are separated by any surface.
  Experiments on ShapeNet show that GIFS outperforms previous state-of-the-art methods in terms of reconstruction quality, rendering efficiency, and visual fidelity.
  Project page is available at \url{https://jianglongye.com/gifs} .
\end{abstract}

\section{Introduction}

\begin{figure}[t]
  \centering
  \includegraphics[width=0.9\columnwidth]{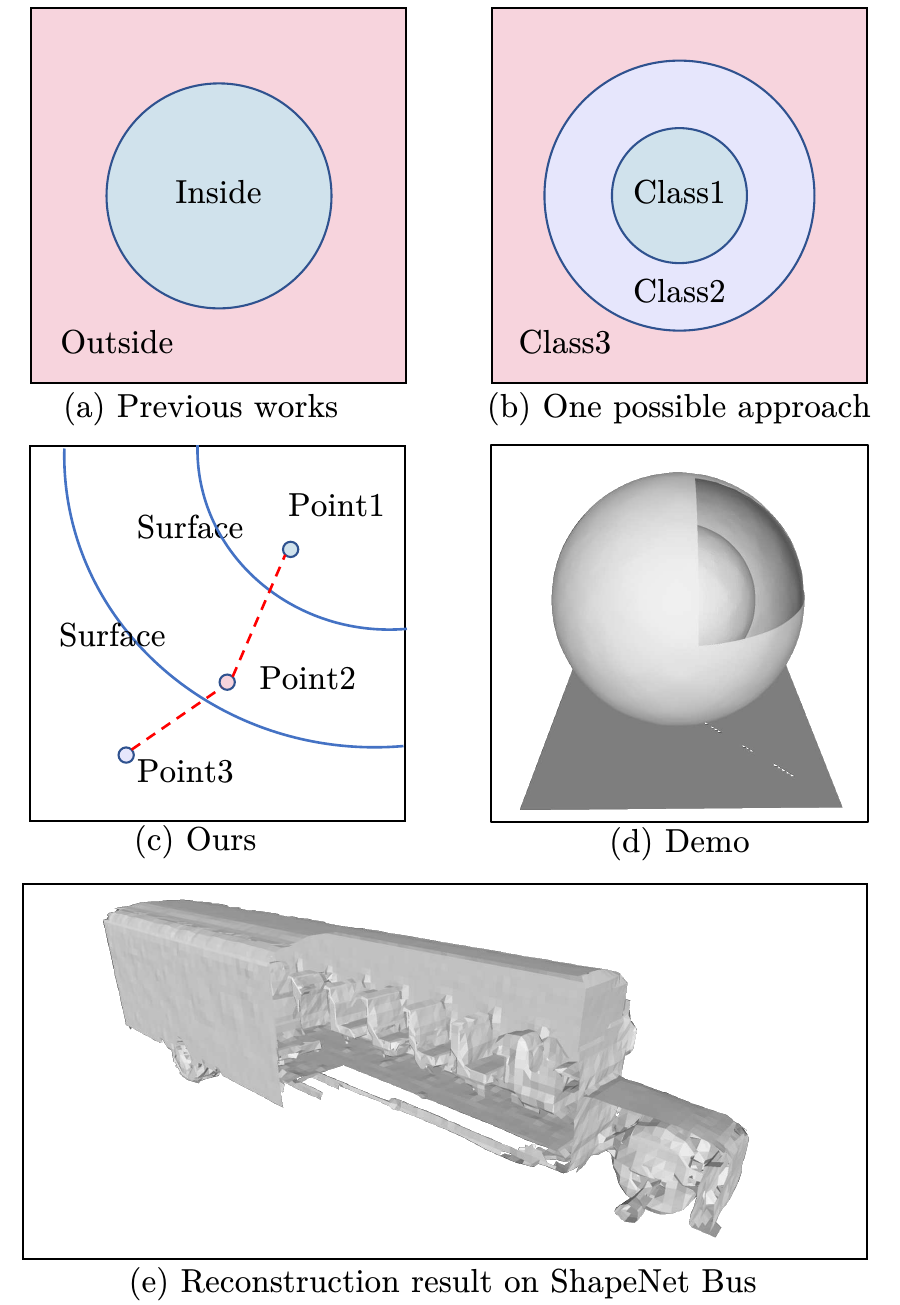}
  \vspace{-0.1in}
  \caption{\textbf{Comparison of different methods.} Previous works divide the space into inside and outside like (a). One possible approach for multi-layer shape is to perform multi-class classification like (b). Our method no longer assigns predefined labels to points but classifies whether two points are separated by any surface, as shown in (c). In (d) and (e), we present a demo shape and a reconstructed bus from ShapeNet respectively. }
  \label{fig: teaser}
  \vspace{-0.15in}
\end{figure}

3D shape reconstruction is a fundamental problem in computer vision, robotics, and graphics. The choice of shape representation heavily affects the reconstruction quality. In recent years, Neural Implicit Functions (NIFs) have achieved high quality and high resolution reconstruction with a compact representation~\cite{chen2019learning, mescheder2019occupancy, park2019deepsdf, chabra2020deep, jiang2020local}. Most NIF approaches employ a neural network to classify whether a 3D point is inside or outside the surface~\cite{mescheder2019occupancy} as shown in Figure~\ref{fig: teaser}(a) or compute the signed distance from a 3D point to one surface~\cite{park2019deepsdf}. However, most shapes in the real world cannot be simply represented by one watertight surface.

Considering multi-layer shapes (e.g., cars with seats inside), binary classification of the 3D space is insufficient.
To tackle this problem, one possible approach is to extend the original definition from binary classification to multi-class classification. We take the double-layer circle as an example and illustrate this idea in Figure~\ref{fig: teaser}(b).
The limitation of the extension is that shapes with different layers require different models.
Therefore it can not represent general shapes in a unified approach.
This method will also fail to represent non-watertight shapes (e.g., garments) since they cannot be partitioned into any number of regions.

In this paper, we introduce a \emph{\textbf{G}eneral \textbf{I}mplicit \textbf{F}unction for 3D \textbf{S}hape} (GIFS). Along with this representation, we further design an algorithm to extract explicit surfaces from GIFS.
Instead of dividing a 3D space into predefined categories like Figure~\ref{fig: teaser}(a) and (b), our representation focuses on the relationships between every two 3D points.
Specifically, we define a binary flag to indicate whether two points are on the same side of object surfaces (see Sec.~\ref{subsec: shape} for formal definition).
As shown in Figure~\ref{fig: teaser}(c), three points are separated by different object surfaces, and our method no longer assigns predefined labels to points but compares between them. We embed 3D points into latent space and employ neural networks to approximate binary flags based on their embeddings. In this way, our method is still able to separate different regions, and the shape is implicitly represented by the decision boundary of the binary flags.

Compared to all previous NIFs that model the relationship between points and surfaces (occupancy~\cite{mescheder2019occupancy}, signed distance~\cite{park2019deepsdf} and unsigned distance~\cite{DBLP:conf/nips/ChibaneMP20}), modeling the relationships between different points with GIFS allows  representing general shapes including multi-layer and non-watertight shapes. To recover the shape from GIFS, we modify the Marching Cubes algorithm to build triangle faces from sampled binary flags. We show a demo shape represented by our method in Figure~\ref{fig: teaser}(d), it is a double-layered ball on top of a non-watertight plane.
The larger ball (outside surface) is cut for better visualization.

Experiments on ShapeNet~\cite{chang2015shapenet} demonstrate that GIFS achieves state-of-the-art performance, and can be trained directly on the raw data.
Directly trainable on raw data not only avoids tedious data processing but also prevents the loss of accuracy during pre-processing.
Compared with NDF~\cite{DBLP:conf/nips/ChibaneMP20}, another generalized shape representation outputs point cloud, our GIFS is far more efficient and visually appealing due to the straightforward mesh generation. We show an example of the reconstruction results on ShapeNet in Figure~\ref{fig: teaser} (e), which is a bus with multiple seats inside.

We highlight our main contributions as follows:
\begin{itemize}
  \vspace{-0.05in}
  \item We introduce a novel generalized shape representation based on spatial relationship between points.
        \vspace{-0.1in}
  \item We design a learning-based implementation GIFS and a surface extraction algorithm for it.
        \vspace{-0.1in}
  \item Our method achieves state-of-the-art performance in 3D reconstruction for watertight and more general shapes and shows advantages in rendering efficiency and visual fidelity over existing approaches.
\end{itemize}

\section{Related Work}

\subsection{Learning Explicit Shape Representations}

Explicit shape representations for 3D shape learning can be roughly classified into three categories: point-based, mesh-based, and voxel-based methods.

\noindent
\textbf{Point-based methods.} As direct outputs of many sensors (e.g., LiDAR, depth cameras), point clouds are a popular representation for 3D learning.
PointNet based architectures~\cite{qi2017pointnet, DBLP:conf/nips/QiYSG17} utilize maxpool operator to keep permutation invariance and are widely used to extract geometry features from point clouds. In addition, recent research for learning with point clouds includes kernel point convolution~\cite{wu2019pointconv, thomas2019kpconv, fan2020pstnet}, graph-based architecture~\cite{landrieu2018large, wang2019dynamic}, transformer~\cite{guo2021pct, Hui2021PyramidPC} and so on.
Point clouds are also served as a output representation in 3D reconstruction~\cite{fan2017point,DBLP:conf/nips/ChibaneMP20}. However, unlike other representations, point clouds do not describe topology and require non-trivial post-processing steps~\cite{bernardini1999ball, kazhdan2006poisson, kazhdan2013screened, calakli2011ssd} to generate renderable surfaces.

\noindent
\textbf{Mesh-based methods.} Mesh consists of a set of vertices and edges, which can be defined as a graph. Hence, graph convolution can be directly applied on mesh for geometry learning~\cite{guo20153d, bronstein2017geometric}. Mesh can also be considered as an output representation in shape reconstruction. Most methods~\cite{wang2018pixel2mesh, lin2019photometric, kanazawa2018learning, ranjan2018generating} deform a template and thereby are limited to a fixed topology.
More recent methods predict vertices and faces directly~\cite{groueix2018papier, dai2019scan2mesh, gkioxari2019mesh}, but do not guarantee surface continuity and are prone to generating self-intersecting faces.
Mesh is also popular in human representations and is utilized to estimate human shape, pose~\cite{kanazawa2018end,kolotouros2019learning,kolotouros2019convolutional}, and garments~\cite{alldieck2019tex2shape, bhatnagar2019multi}. The underlying mesh model~\cite{loper2015smpl, pons2017clothcap} still restricts the topology and details.

\noindent
\textbf{Voxel-based methods.} Voxels are an intuitive 3D extension of pixels in 2D images. The common learning paradigm (i.e., convolution) in the 2D domain could be extended to voxels naturally. Consequently, voxels are widely used for shape learning~\cite{DBLP:conf/nips/KarHM17, ji2017surfacenet, jimenez2016unsupervised}.
The occupancy grid is the simplest use case of voxel representation.
However, due to the cubically growing memory footprint, the resolution of the grid is practically limited to a certain level~\cite{liao2018deep, wu20153d,choy20163d}.
Therefore, the voxel-based methods struggle to reconstruct high-fidelity shapes with fine details. Higher resolutions can be achieved at the cost of limited training batches and slow training~\cite{wu2016learning,DBLP:conf/nips/0001WXSFT17}, or complex multi-resolution implementations~\cite{hane2017hierarchical,tatarchenko2017octree}. Replacing occupancy with Truncated Signed Distance functions~\cite{curless1996volumetric} for learning~\cite{dai2017shape, ladicky2017point} allows for sub-voxel precision, although the resolution is still limited by the fixed grid.

\subsection{Learning Implicit Shape Representations}

In the past few years, a lot of advances have been made in 3D shape learning with neural implicit function~\cite{park2019deepsdf, mescheder2019occupancy, chen2019learning, gropp2020implicit, xu2019disn, saito2019pifu, saito2020pifuhd, DBLP:conf/cvpr/GenovaCSSF20}. Despite the differences in output (occupancy~\cite{mescheder2019occupancy, saito2019pifu}, signed distance~\cite{park2019deepsdf, genova2019learning}, or unsigned distance~\cite{DBLP:conf/nips/ChibaneMP20}), they all learn a continuous function to predict the relationship between the querying point (x-y-z) and surface.
Pioneering methods~\cite{park2019deepsdf, mescheder2019occupancy, chen2019learning} represent shape with a global latent code. An encoder~\cite{mescheder2019occupancy, chen2019learning} or an optimization method~\cite{park2019deepsdf} is adopted to embed the shape. And then a decoder is used to reconstruct shape from the latent code. For constructing a smooth latent space, some techniques such as curriculum learning~\cite{duan2020curriculum}, and adversarial training~\cite{kleineberg2020adversarial} are exploited. Following works~\cite{sitzmann2020implicit,DBLP:conf/nips/TancikSMFRSRBN20} also use periodic activation functions to map the input position to high dimensional space to preserve high-frequency details. Unlike the global methods, the local methods divide the 3D space into uniform grids~\cite{jiang2020local, chabra2020deep,chibane2020implicit} or local parts~\cite{genova2019learning, DBLP:conf/cvpr/GenovaCSSF20}, and the latent code varies by local grids/parts. Since each latent code only needs to represent the shape of a local area, better details can be preserved and generalizability is improved.

The methods mentioned above rely on inside and outside partition and thus always require tedious data processing to artificially close the shapes, which results in non-trivial loss of details and inner structures.
Recent works~\cite{atzmon2020sal, zhao2021sign, atzmon2020sald} introduce sign agnostic learning to learn from the raw data. However, the output is again signed distance and hence can not reconstruct general shapes.
To model general shapes, NDF~\cite{DBLP:conf/nips/ChibaneMP20} proposes to predict unsigned distance fields (UDFs) to represent general shapes. However, the output of NDF is point cloud, which does not only rely on an expensive post-processing step but also struggles to obtain high-quality final meshes. Following works~\cite{venkatesh2021deep, zhao2021learning} incorporate normals and gradients to improve reconstruction accuracy and efficiency.

In contrast, we propose to predict whether two different points are separated by any surface, and it allows general shape representation. To the best of our knowledge, our method is the first to utilize a neural network to model the relationship between different points in shape representation. Since the different regions are separated in our representation, a Marching Cubes-like algorithm can be employed to directly convert our representation into a mesh rather than a point cloud. Due to this, we outperform NDF in terms of efficiency and visual effect.

\section{Method}

\subsection{Generalized Shape Representation}
\label{subsec: shape}

Instead of modeling the relationship between points and surfaces such as~\cite{mescheder2019occupancy, saito2019pifu, park2019deepsdf, genova2019learning, DBLP:conf/nips/ChibaneMP20}, we propose a novel shape representation focusing on the relationship between different points. We utilize a binary flag to indicate whether two points are separated by any surface. Specifically, the pair of points are considered to be separated if their line segment intersects any surface.

\begin{figure}[t]
  \centering
  \includegraphics[width=1.0\columnwidth]{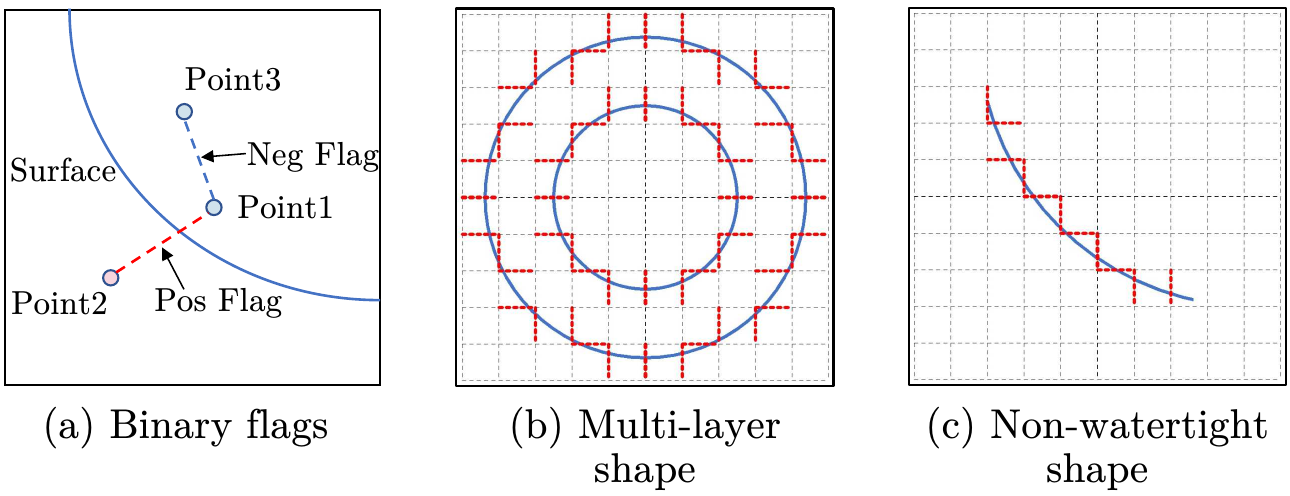}
  \vspace{-0.2in}
  \caption{\textbf{Generalized shape representation.} In (a), we show examples of the binary flag. The flag of the points separated by the surface is positive, otherwise, it is negative. In (b) and (c), we show how a double-layer circle and a non-watertight curve are represented. The positive flags are bolded red. }
  \vspace{-0.2in}
  \label{fig: representation}
\end{figure}

\begin{figure*}[t]
  \centering
  \includegraphics[width=0.95\textwidth]{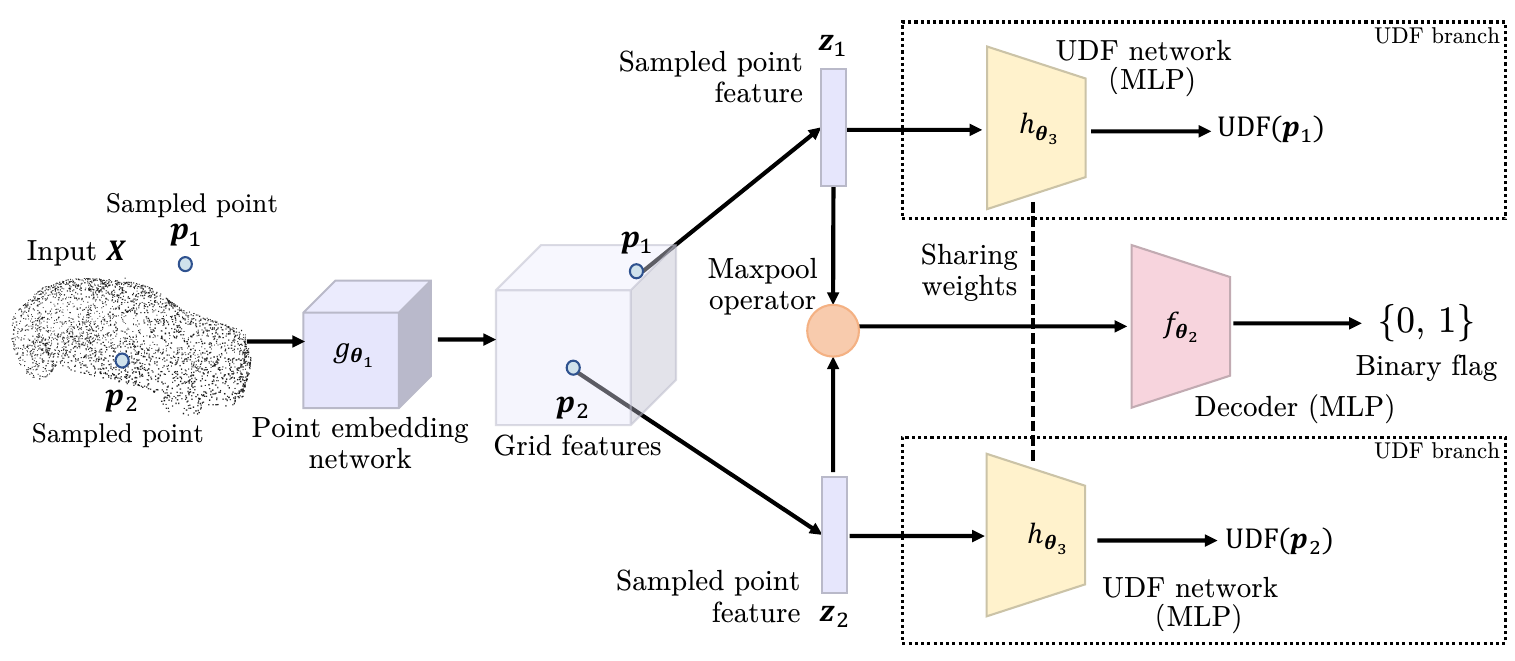}
  \vspace{-0.1in}
  \caption{\textbf{Overview of GIFS.} We learn a neural implicit function to classify whether two points are separated by any surface. First, the input point cloud is encoded to a grid of features. Given two 3D points, corresponding features are extracted from the grid. Then a maxpool operator is utilized to keep the permutation invariance. The decoder takes the fused feature as input and approximates the binary flag between points. An extra UDF branch can be used to enhance the spatial perception of the feature. }
  \label{fig: method}
  \vspace{-0.1in}
\end{figure*}

Let the surface of object be a subset $\boldsymbol{S} \subset \mathbb{R}^{3}
$. Given two points $\boldsymbol{p}_1 \in \mathbb{R}^3$ and $\boldsymbol{p}_2 \in \mathbb{R}^3$, we can get a segment $\boldsymbol{e}$ formed by these two points:  $\boldsymbol{e}(\boldsymbol{p}_1, \boldsymbol{p}_2) = \{ \boldsymbol{p}_1 + k * (\boldsymbol{p}_2 - \boldsymbol{p}_1) \,| \, k \in [0, 1] \}$. If the segment intersects the surface, there is at least one point that belongs to both the segment and the surface. The binary flag $\boldsymbol{b}$ used to indicate whether the segment intersects the surface is defined as:
\begin{equation}
  \boldsymbol{b}(\boldsymbol{p}_1, \boldsymbol{p}_2, \boldsymbol{S}) =
  \begin{cases}1, & {\exists \boldsymbol{p} \in \boldsymbol{e}(\boldsymbol{p}_1, \boldsymbol{p}_2), \boldsymbol{p} \in \boldsymbol{S}} \\ 0,  & \text{otherwise} \end{cases}
\end{equation}
Figure~\ref{fig: representation} (a) shows the example of the binary flags: the flag of Point 1 and Point 2 which are separated by any surface is positive (1), and the flag of Point 1 and Point 3 is negative (0).

The definition differs from the connected components in computer vision. Our binary flag is not transitive in 3D space, i.e., $\boldsymbol{b}(\boldsymbol{p}_1, \boldsymbol{p}_2, \boldsymbol{S}) = \boldsymbol{b}(\boldsymbol{p}_2, \boldsymbol{p}_3, \boldsymbol{S})$ does not imply that $\boldsymbol{b}(\boldsymbol{p}_1, \boldsymbol{p}_2, \boldsymbol{S}) = \boldsymbol{b}(\boldsymbol{p}_2, \boldsymbol{p}_3, \boldsymbol{S}) = \boldsymbol{b}(\boldsymbol{p}_1, \boldsymbol{p}_3, \boldsymbol{S})$. This is the reason that our definition allows to represent non-watertight shapes.

Similar to the previous implicit function, the shape is still represented by the decision boundary. Different regions are separated, and the binary flag changes when crossing the surface. Figure~\ref{fig: representation} (b) and (c) show how a double-layer circle and a non-watertight curve are represented. The flags of the points crossing the circle and the curve are positive (bolded red in the figure), and other flags are negative (gray in the figure). Note that the input points of the binary flag are not limited to horizontal or vertical points, and the resolution will be much higher when extracting surfaces.

\subsection{Learning General Implicit Function for 3D Shape}
\label{subsec: gif}

We propose to learn a \emph{\textbf{G}eneral \textbf{I}mplicit \textbf{F}unction for 3D \textbf{S}hape} (GIFS) to estimate the binary flag $\boldsymbol{b}$. It takes two 3D points $\boldsymbol{p}_1$ and $\boldsymbol{p}_2$ plus an observation $\boldsymbol{X}$ as inputs.  In this paper, we study the task of shape reconstruction from sparse point cloud $\boldsymbol{X} \in \mathbb{R}^{n \times 3}$. For completeness, we first explain the encoding module, which follows IF-Net~\cite{chibane2020implicit}. Then we describe our decoding module and learning procedure. The overview of our method is shown in Figure~\ref{fig: method}.

\noindent
\textbf{Point embedding.} The input point cloud is first converted to a discrete voxel grid. Then a 3D CNN is applied to get the multi-scale grid features $\boldsymbol{F}_{1}, . ., \boldsymbol{F}_{n}, \boldsymbol{F}_{k} \in \mathcal{F}_{k}^{K \times K \times K}$, where $K$ is the grid size varies with scale and $\mathcal{F}_{k} \in \mathbb{R}^{C}$ is a deep feature with channels $C$. Refer to IF-Net~\cite{chibane2020implicit} for further details. Given a query point $\boldsymbol{p}$, a set of deep features is sampled from the multi-scale encoding and is concatenated as the embedding $\boldsymbol{z} \in \mathcal{F}_{1} \times \ldots \times \mathcal{F}_{n}$ of the point. We denote the point embedding network as:
\begin{equation}
  \boldsymbol{z} = g_{\boldsymbol{\theta}_1}(\boldsymbol{X}, \boldsymbol{p}) \colon \mathbb{R}^n \times \mathbb{R}^3 \mapsto \mathcal{F}_{1} \times \ldots \times \mathcal{F}_{n},
\end{equation}
where $\boldsymbol{\theta}_1$ is the learnable parameters for embedding network. The purple cube in Figure~\ref{fig: method} indicates the embedding network.

\noindent
\textbf{Binary flag prediction.} A decoder is implemented to predict the binary flag of points. To keep the permutation invariance, a maxpool operator is applied to the points embeddings $\boldsymbol{z}_1$ and $\boldsymbol{z}_2$: $\boldsymbol{z}_f = \texttt{MAX} (\boldsymbol{z}_1, \boldsymbol{z}_2)$, where $\boldsymbol{z}_f$ is the fused feature of the two points. Then the fused feature $\boldsymbol{z}_f$ is passed to the decoder to predict the binary flag $\boldsymbol{b}$. The decoder $f$ is a multilayer perceptron parameterized by $\boldsymbol{\theta}_2$:
\begin{equation}
  f_{\boldsymbol{\theta}_2}(\boldsymbol{z}_f) \approx \boldsymbol{b} \colon \mathcal{F}_{1} \times \ldots \times \mathcal{F}_{n} \mapsto [0, 1],
\end{equation}
The pink block in Figure~\ref{fig: method} indicates the decoder.

\noindent
\textbf{UDF branch.} Although the above decoder is sufficient for shape reconstruction, we additionally incorporate a UDF branch to enhance the spatial perception of the point feature and accelerate the surface extraction (see Sec.~\ref{subsec: extraction}). The unsigned distance function (UDF) is defined as the distance between the querying point and its closet point on the object surface. Therefore it does not introduce the traditional inside/outside partition. NDF~\cite{DBLP:conf/nips/ChibaneMP20} is the first to use neural implicit function to approximate UDF. Given a embedding $\boldsymbol{z}$ of point $\boldsymbol{p}$ , the UDF network $h$ is denoted as:
\begin{equation}
  h_{\boldsymbol{\theta}_3}(\boldsymbol{z}) \approx \texttt{UDF}(\boldsymbol{p}) \colon \mathcal{F}_{1} \times \ldots \times \mathcal{F}_{n} \mapsto \mathbb{R}_0^{+},
\end{equation}
where $\boldsymbol{\theta}_3$ is the learnable parameters for the UDF network and $\texttt{UDF}(\boldsymbol{p})$ is the ground truth UDF for point $\boldsymbol{p}$. The yellow block in Figure~\ref{fig: method} indicates the UDF network. Ablation studies on the UDF branch are shown in Sec.~\ref{subsec: ablation}.

\noindent
\textbf{Learning procedure.} To train the point embedding network $ g_{\boldsymbol{\theta}_1}(\cdot)$, the decoder $f_{\boldsymbol{\theta}_2}(\cdot)$ and the UDF network $h_{\boldsymbol{\theta}_3}(\cdot)$, pairs $\left\{\boldsymbol{X}_{i}, \mathcal{S}_{i} | i \in 1, \ldots, N \right\}$ of observation $\boldsymbol{X}_{i}$ and corresponding ground truth shape $\mathcal{S}_{i}$ are required, where $i$ is the index of observation and $N$ is the number of training examples. For each training example, we sample a pair $\left\{\boldsymbol{p}_1, \boldsymbol{p}_2 \right\}$ of two points and compute their ground truth $\texttt{UDF}(\boldsymbol{p}, \mathcal{S})$ as well as binary flags $\boldsymbol{b}(\boldsymbol{p}_1, \boldsymbol{p}_2, \mathcal{S})$. During training, two loss terms are exploited: $\mathcal{L}_{b}$ for binary flag prediction and $\mathcal{L}_{u}$ for UDF regression. The $\mathcal{L}_{b}$ is defined as
\begin{multline}
  \mathcal{L}_{b} (\boldsymbol{X}, \boldsymbol{p}_1, \boldsymbol{p}_2) = \\ \left| f_{\boldsymbol{\theta}_2}(\texttt{MAX}(g_{\boldsymbol{\theta}_1}(\boldsymbol{X}, \boldsymbol{p}_1),  g_{\boldsymbol{\theta}_1}(\boldsymbol{X}, \boldsymbol{p}_2))) - \boldsymbol{b}(\boldsymbol{p}_1, \boldsymbol{p}_2, \mathcal{S}_{\boldsymbol{X}}) \right|,
\end{multline}
where $\mathcal{S}_{\boldsymbol{X}}$ is the corresponding ground truth shape for observation $\boldsymbol{X}$.
And the $\mathcal{L}_{u}$ is defined as:
\begin{multline}
  \mathcal{L}_{u} (\boldsymbol{X}, \boldsymbol{p}) = \\ \left|\min \left(h_{\boldsymbol{\theta}_3}(g_{\boldsymbol{\theta}_1}(\boldsymbol{X}, \boldsymbol{p})), \delta\right)-\min \left(\texttt{UDF}\left(\boldsymbol{p}, \mathcal{S}_{\boldsymbol{X}}\right), \delta\right)\right|,
  \label{eq: udf_loss}
\end{multline}
where $\delta > 0$ is a threshold which concentrates the model capacity to represent the near space of the surface.

The parameters are optimized by minimizing the following mini-batch loss:
\begin{multline}
  \mathcal{L}_{\mathcal{B}}(\boldsymbol{\theta}_1, \boldsymbol{\theta}_2, \boldsymbol{\theta}_3)= \sum_{\boldsymbol{X} \in \mathcal{B}} \sum_{\boldsymbol{p}_1, \boldsymbol{p}_2 \in \mathcal{P}} \mathcal{L}_{b}(\boldsymbol{X}, \boldsymbol{p}_1, \boldsymbol{p}_2) + \\
  \lambda (\mathcal{L}_{u}(\boldsymbol{X}, \boldsymbol{p}_1) + \mathcal{L}_{u}(\boldsymbol{X}, \boldsymbol{p}_2)),
  \label{eq: overall_loss}
\end{multline}
where $\mathcal{B}$ is a mini-batch of input point cloud observations, $\mathcal{P}$ is a subset of sampled pairs for training and $\lambda$ is the weight for $\mathcal{L}_{u}$.

\subsection{Surface Extraction from GIFS}
\label{subsec: extraction}

Since the proposed method is different from previous NIFs, it's non-trivial to design an algorithm for surface extraction. We adapt Marching Cubes~\cite{lorensen1987marching} algorithm to extract explicit surface (mesh) from our implicit representation. Marching Cubes divides the space into the 3D grid and locates cubes that intersect the object's surface. For the intersected cube, triangles are created based on the occupancies of 8 vertices. A lookup table consisting of $2^8 = 256$ different assignments is used to increase the speed.

Although there is no occupancy in our method, we make an assumption that on a micro-scale, the space near the surface can always be divided into two classes. Based on that, we divide the space into the 3D grid and perform binary classification on the 8 vertices in each cell according to the binary flags between them (generated by our method). Then we utilize the lookup table in Marching Cubes to create triangles. Let $\boldsymbol{c}_i \in \{0, 1\}$ be the assignment for vertex $i \in \{0, 1 \ldots 7\}$ and $\boldsymbol{b}_{ij} \in [0, 1]$ be the flag between vertex $i$ and vertex $j$. The cost function for computing the assignment $\boldsymbol{c}_i$ is:
\begin{equation}
  \mathcal{L} = \sum_{i}\sum_{j \neq i} \left| \boldsymbol{c}_i -  \boldsymbol{c}_j \right|  \left| 1 - \boldsymbol{b}_{ij} \right|.
\end{equation}
The optimal assignment can be found by minimizing the cost function. We simply apply a brute-force algorithm in implementation since the speed is still acceptable.

\begin{figure*}[t]
  \centering
  \includegraphics[width=1.0\textwidth]{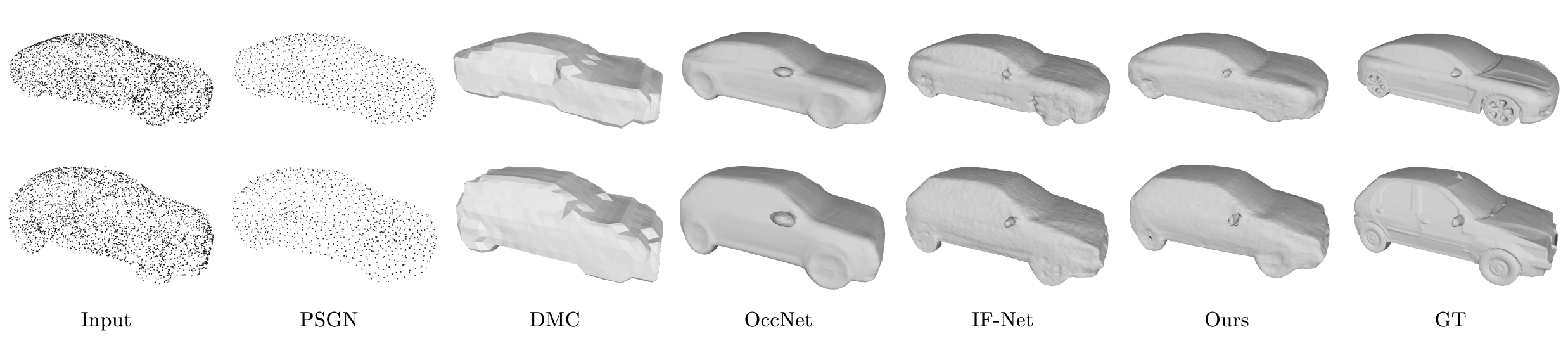}
  \vspace{-0.2in}
  \caption{\textbf{Comparison of methods trained on watertight shapes.} Our method can reconstruct watertight shapes with the same accuracy as the state-of-the-art method.}
  \label{fig: close0}
  \vspace{-0.1in}
\end{figure*}

\noindent
\textbf{Coarse-to-fine surface extraction.} Similar to OccNet~\cite{mescheder2019occupancy}, a coarse-to-fine paradigm is adopted to speed up the extraction process. We start from a low-resolution grid and evaluate the UDF at the center of each cube. If the UDF value is smaller than the cube size multiples some threshold $\tau$, we consider the cube intersects the surface and subdivide it into 8 subcubes. The evaluation is repeated on the new cubes until the desired final resolution is reached. We perform our adapted Marching Cubes on final cubes.

\noindent
\textbf{Mesh refinement.} The initial mesh extracted by adapted Marching Cubes is only an approximation of the decision boundary and can be further refined with the UDF value. We randomly sample $N$ points $\boldsymbol{p}_i$ from each face of the output mesh and refine it by minimizing the loss:
\begin{equation}
  \mathcal{L} = \sum_{i}^{N} \left|h_{\boldsymbol{\theta}_3}(g_{\boldsymbol{\theta}_1}(\boldsymbol{X}, \boldsymbol{p}_i))\right|.
\end{equation}
We further push the mesh surface to the decision boundary by minimizing the UDF value of the surface. We report an ablation study for the mesh refinement in Sec.~\ref{subsec: ablation}.

\section{Experiments}

We focus on the task of 3D shape reconstruction from sparse point clouds to validate GIFS. We first show that GIFS can reconstruct watertight shapes on par with the state-of-the-art methods in Sec.~\ref{subsec: closed}, and then show GIFS can reconstruct general shapes including multi-layer, non-watertight shapes in Sec.~\ref{subsec: general}. Ablation studies on architecture, grid size and sampling strategy are shown in Sec.~\ref{subsec: ablation}.

\subsection{Experimental Settings}
\label{subsec: setting}

\begin{figure*}[t]
  \centering
  \includegraphics[width=0.95\textwidth]{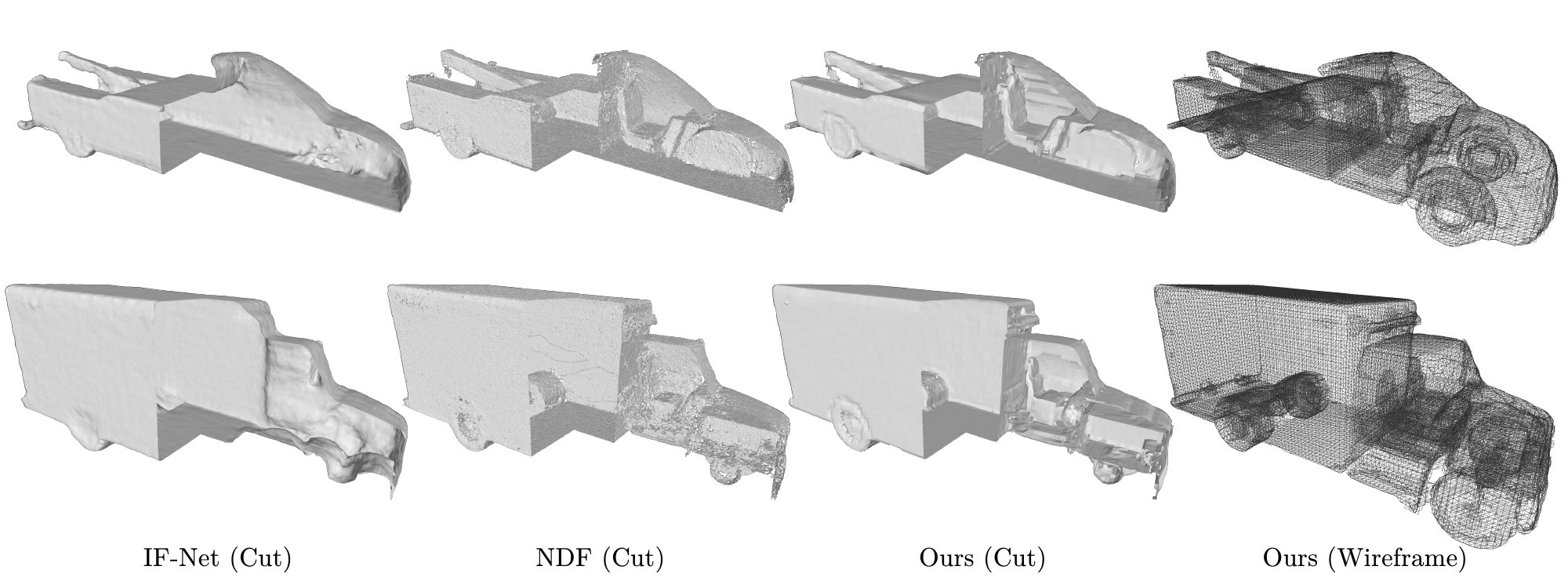}
  \vspace{-0.1in}
  \caption{\textbf{Reconstruction results of different methods on general shapes.} Our method can reconstruct smooth, continuous surfaces, and achieve a better visual effect than previous methods. }
  \label{fig: general0}
\end{figure*}

\noindent
\textbf{Datasets.} In our experiments, we follow the setting of NDF and choose the ``Car" category of the ShapeNet~\cite{chang2015shapenet} dataset, which consists of 7498 synthetic objects. The ``Car" category has the largest amount of multi-layer shapes as well as non-watertight shapes. For watertight shapes, we use the common training/test split from 3D-R2N2~\cite{choy20163d} and the processed watertight meshes from DISN~\cite{xu2019disn}. For general shapes, we use the training/test split from NDF~\cite{DBLP:conf/nips/ChibaneMP20}. During training and evaluation, all meshes are normalized and centered to a unit cube. Besides, we employ the MGN~\cite{bhatnagar2019multi} dataset, which consists of 307 garments, to show the representation power of our method.

\noindent
\textbf{Evaluation Metrics.} To measure the reconstruction quality, we adopt established metrics defined in LDIF~\cite{DBLP:conf/cvpr/GenovaCSSF20}: the Chamfer distance and F-Score. The Chamfer distance measures the average errors of all points and is sensitive to outliers. F-Score measures the percentage of good predictions. For evaluation, we randomly sample 100k points from the surface to calculate all metrics. The thresholds for F-Score are 0.01 (same as LDIF) and 0.005 (for better comparison).  Lower is better for Chamfer distance; higher is better for F-Score.

\noindent
\textbf{Baselines.} For the experiment of watertight shapes, we compare our method to Point Set Generation Networks~\cite{fan2017point} (PSGN), Deep Marching Cubes~\cite{liao2018deep} (DMC), Occupancy Network~\cite{mescheder2019occupancy} (OccNet), and IF-Net~\cite{chibane2020implicit}, the state-of-the-art in shape reconstruction. For the experiment of general shapes, we compare our method to NDF~\cite{DBLP:conf/nips/ChibaneMP20}. For methods that the authors do not provide a pretrained model, we retrain it until the minimum validation error is reached.

\noindent
\textbf{Implementation details.} The resolution of the voxel grid in the point embedding network is $128^3$ for 3000 points input and $256^3$ for 10000 points input. The decoder for binary flag prediction is a 5-layer multilayer perceptron. All internal layers are 256-dimensional and have ReLU non-linearities. The UDF network is exactly the same as the decoder in NDF. The $\delta$ in Eq.~\ref{eq: udf_loss} is set to 0.1, the $\lambda$ in Eq.~\ref{eq: overall_loss} is set to 10. During training, we employ the Adam optimizer with a learning rate of $1 \times 10^{-4} $. During inference, the initial resolution of the grid is $20^3$ and is subdivided 3 times. The final resolution is $160^3$. The $\tau$ in Sec.~\ref{subsec: extraction} is set to 2.

During data generation, we mainly follow the same strategy as NDF~\cite{DBLP:conf/nips/ChibaneMP20}. Specifically, we first sample points $\boldsymbol{p}_{i}^{\mathcal{S}} \in \mathbb{R}^{3}$ on the ground truth surfaces and add two random displacements $\{\boldsymbol{n}^0_{i}, \boldsymbol{n}^1_{i}\} \sim \mathcal{N}(0, \boldsymbol{\Sigma})$ to produce pairs of points $\{\boldsymbol{p}^0_{i}, \boldsymbol{p}^1_{i}\}$, i.e., $\boldsymbol{p}^0_{i} = \boldsymbol{p}_{i}^{\mathcal{S}} + \boldsymbol{n}^0_{i}, \boldsymbol{p}^1_{i} = \boldsymbol{p}_{i}^{\mathcal{S}} + \boldsymbol{n}^1_{i}$, where $\boldsymbol{\Sigma} \in \mathbb{R}^{3 \times 3}$ is a diagonal covariance matrix with entries $\boldsymbol{\Sigma}_{i, i} = \sigma$. We adopt 3 different $\sigma$ of 0.005, 0.01, 0.03.
We also randomly sample 10\% of the data in the 3D grid. We report performances with different combinations of $\sigma$ in Sec.~\ref{subsec: ablation}.
We employ the intersection algorithm from CGAL~\cite{cgal:atw-aabb-21b} to calculate the ground truth binary flags.

\begin{table}[t]
  \centering
  \begin{tabular}{l|cc|cc}
    & \multicolumn{2}{c|}{Chamfer distance $\downarrow$} & \multicolumn{2}{c}{F-Score $\uparrow$} \\
    Method & Mean  & Median  & F1\textsuperscript{0.005} & F1\textsuperscript{0.01} \\
    \shline
    Input & 0.782 & 0.754 & 26.15 & 65.00 \\
    PSGN~\cite{fan2017point} & 2.251 & 2.018 & 4.33 & 24.34 \\
    DMC~\cite{liao2018deep} & 5.963 & 3.654 & 32.22 & 58.06 \\
    OccNet~\cite{mescheder2019occupancy} & 3.251 & 2.386 & 28.75 & 65.41 \\
    IF-Net~\cite{chibane2020implicit} & 0.260 & 0.125 & 86.70 & 96.50 \\
    NDF~\cite{DBLP:conf/nips/ChibaneMP20} & 0.152 & 0.125 & 85.70 & \textbf{98.32} \\
    Ours & \textbf{0.146} & \textbf{0.114} & \textbf{88.75} & 98.09
  \end{tabular}
  \caption{\textbf{Quantitative evaluation on watertight shapes}. We train and evaluate our method on the watertight data of the ShapeNet ``Car" category. We report the mean and median value of the Chamfer distance, as well as F-Score at different thresholds. The Chamfer distance results $\times 10 ^ {-4}$. Our method outperforms baselines in all but one metric. }
  \label{tab: closed}
  \vspace{-0.1in}
\end{table}

\subsection{Shape Reconstruction of Watertight Shapes}
\label{subsec: closed}

We first show the representation power of our method on watertight shapes. We train and evaluate with the processed data provided by DISN. Their data processing step closes shape and removes all interior structures. In this experiment, the input for all methods is 3000 points sampled from the watertight data of the ShapeNet ``Car" category. For a fair comparison, the resolution of the encoding voxel grid in both IF-Net and our method is $128^3$.

Figure~\ref{fig: close0} shows that our method can reconstruct a watertight shape with the same accuracy as the state-of-the-art method. Note that OccNet uses a global code to represent the whole shape and outputs a smoother result which can be wrong. As shown in the figure, the rear-view car mirrors of the OccNet results differ significantly from the GT shape.

We also quantitatively compare our method to baselines in Table~\ref{tab: closed} and show that GIFS outperforms baselines in all but one metric.

\subsection{Shape Reconstruction of General Shapes}
\label{subsec: general}

We show the reconstruction results of our method on general shapes including multi-layer, and non-watertight shapes. Since our method does not require a pre-processing step (e.g., deep fusion), it can directly train and evaluate with the raw ShapeNet data. We also do not lose accuracy due to this step. In this experiment, we mainly compare our method to the NDF, the previous method aimed at modeling general shapes. For a fair comparison, the resolution of the encoding voxel grid in both NDF and our method is $256^3$. The input is 10000 points sampled from the raw data of the ShapeNet ``Car" category.

\begin{figure}[t]
  \centering
  \includegraphics[width=0.95\columnwidth]{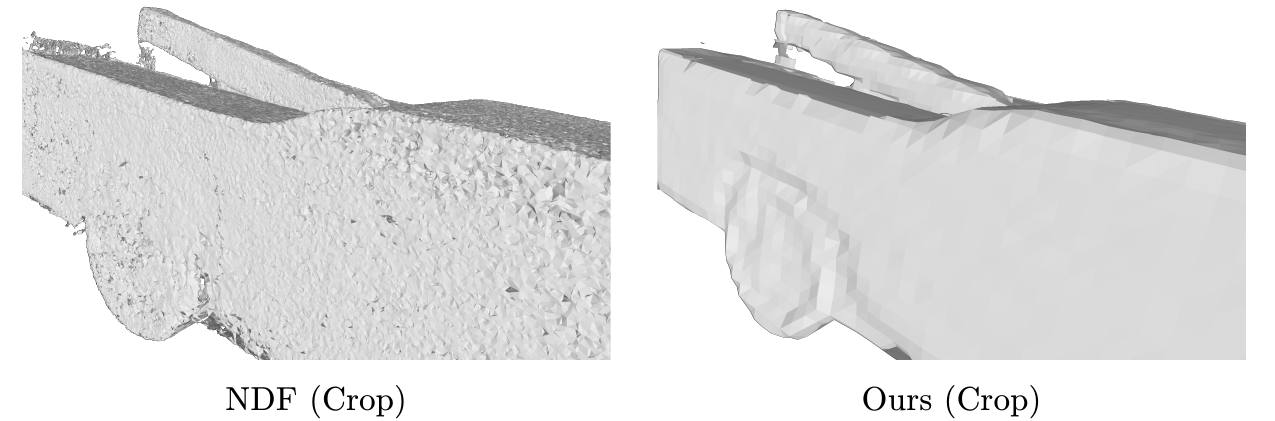}
  \vspace{-0.1in}
  \caption{\textbf{Zoom-in comparison of different methods on general shapes.} The surface in the reconstruction result of NDF is far from smooth and contains holes. }
  \label{fig: general0_crop}
\end{figure}

\begin{figure}[t]
  \centering
  \includegraphics[width=1.0\columnwidth]{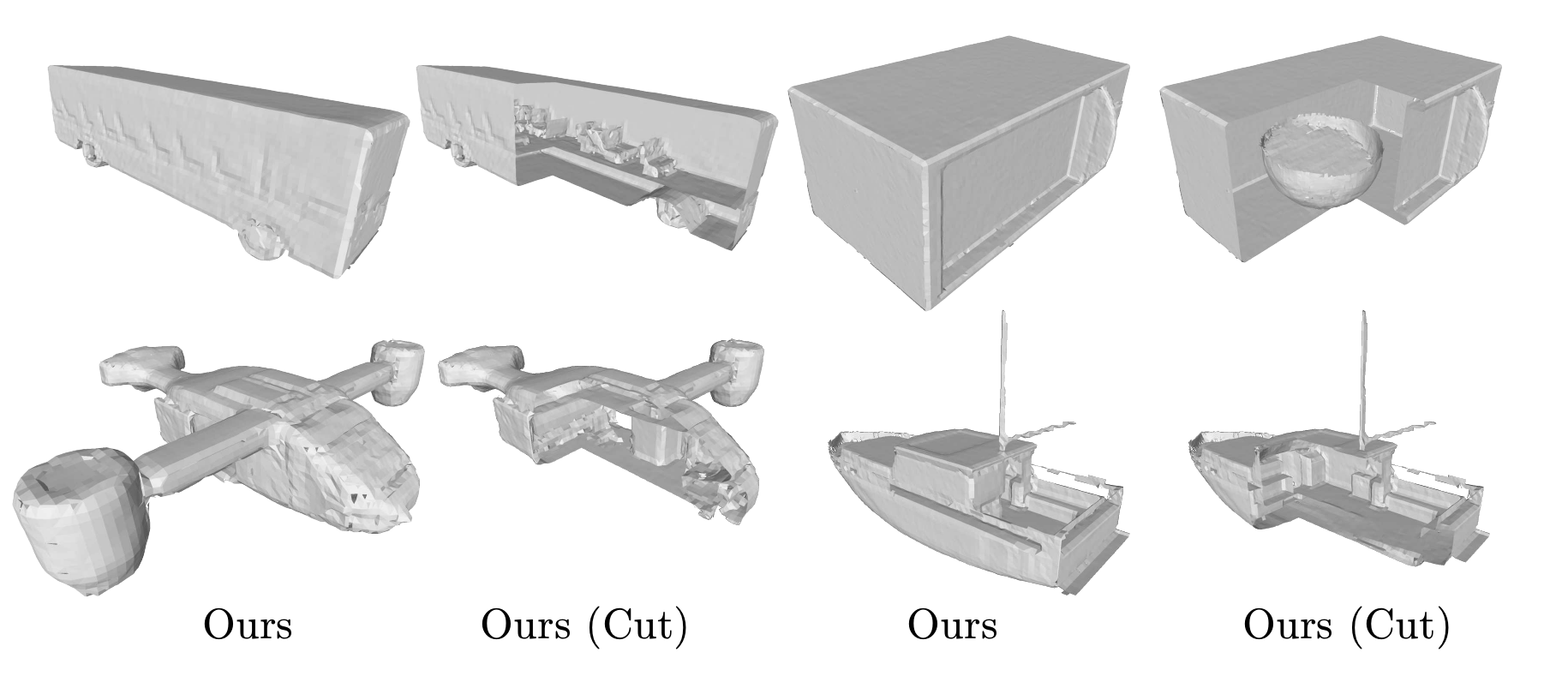}
  \vspace{-0.25in}
  \caption{\textbf{Reconstruction results of multi-layer shapes on other classes.} Our method can reconstruct internal structures of various shapes. }
  \label{fig: general1}
  \vspace{-0.1in}
\end{figure}

\noindent
\textbf{Qualitative analysis.} Figure~\ref{fig: general0} shows reconstruction results of two multi-layer cars from the test set. Traditional implicit neural representations are limited to single-layer, watertight shapes, so all the internal structure of the car is lost in the IF-Net results. NDF can represent general shapes, but its network output is a point cloud and relies on the ball-pivoting algorithm to obtain mesh. Even with carefully selected thresholds, the surface in the reconstruction result of NDF is still far from smooth and has a number of holes. In contrast, our method is able to reconstruct a flat, continuous surface, and thereby achieve a better visual effect. This can be observed more clearly in the zoom-in comparison in Figure~\ref{fig: general0_crop}. Note that the NDF visualization code also includes a filter to close holes, while our method does not apply any filter.

\begin{figure}[t]
  \centering
  \includegraphics[width=1.0\columnwidth]{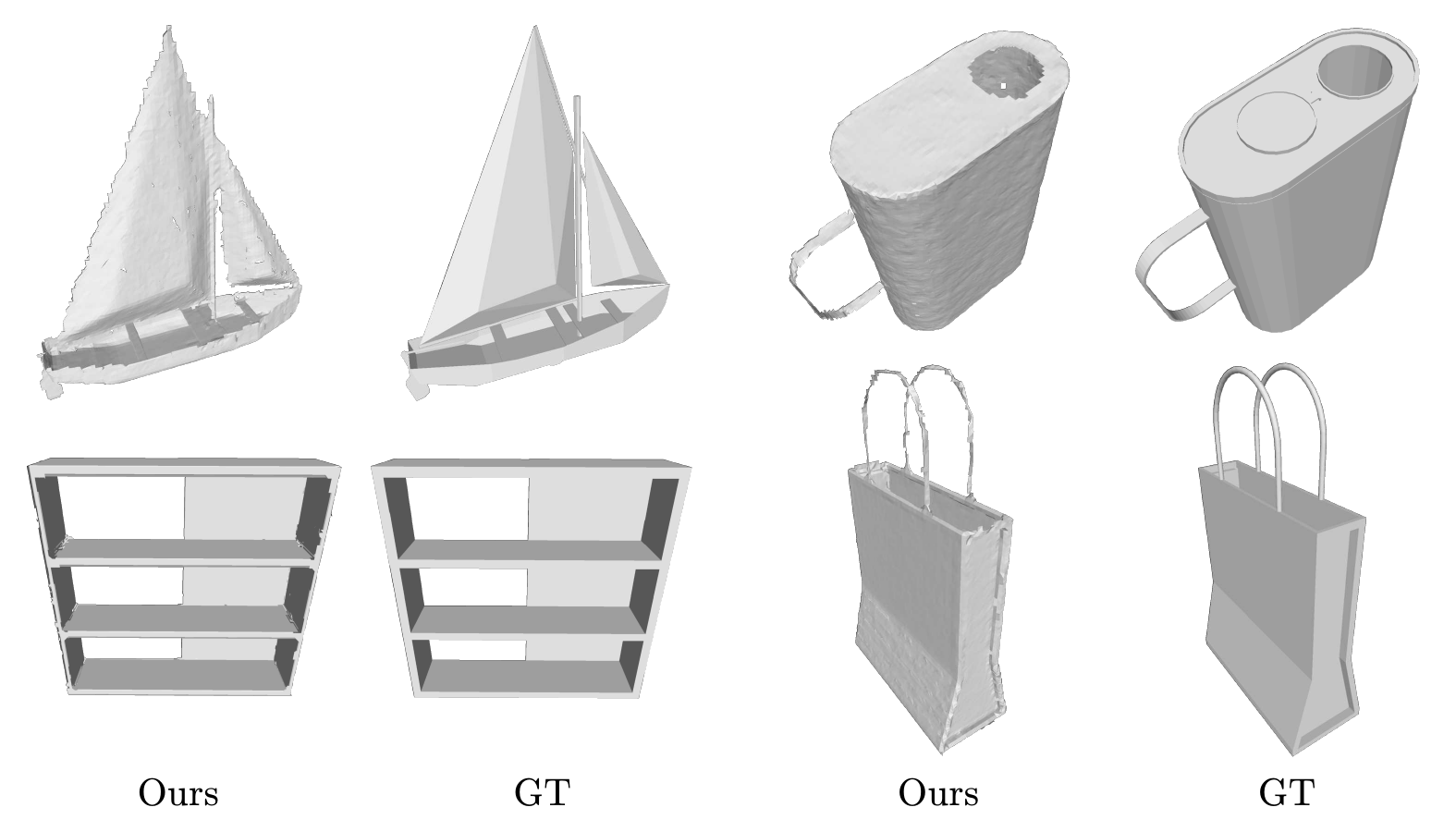}
  \vspace{-0.25in}
  \caption{\textbf{Reconstruction results of non-watertight shapes on other classes.} The non-watertight shapes are difficult for traditional implicit neural representations to reconstruct. }
  \label{fig: general2}
  \vspace{-0.1in}
\end{figure}

Figure~\ref{fig: general1} shows the reconstruction results of multi-layer shapes on other classes. Our method can reconstruct the chairs in the bus and the airplane or the bowl in the microwave. Figure~\ref{fig: general2} shows the reconstruction results of non-watertight shapes. Our method can represent cans with a hole and bag, which are difficult for the traditional implicit function to represent.

\begin{figure}[t]
  \centering
  \includegraphics[width=0.95\columnwidth]{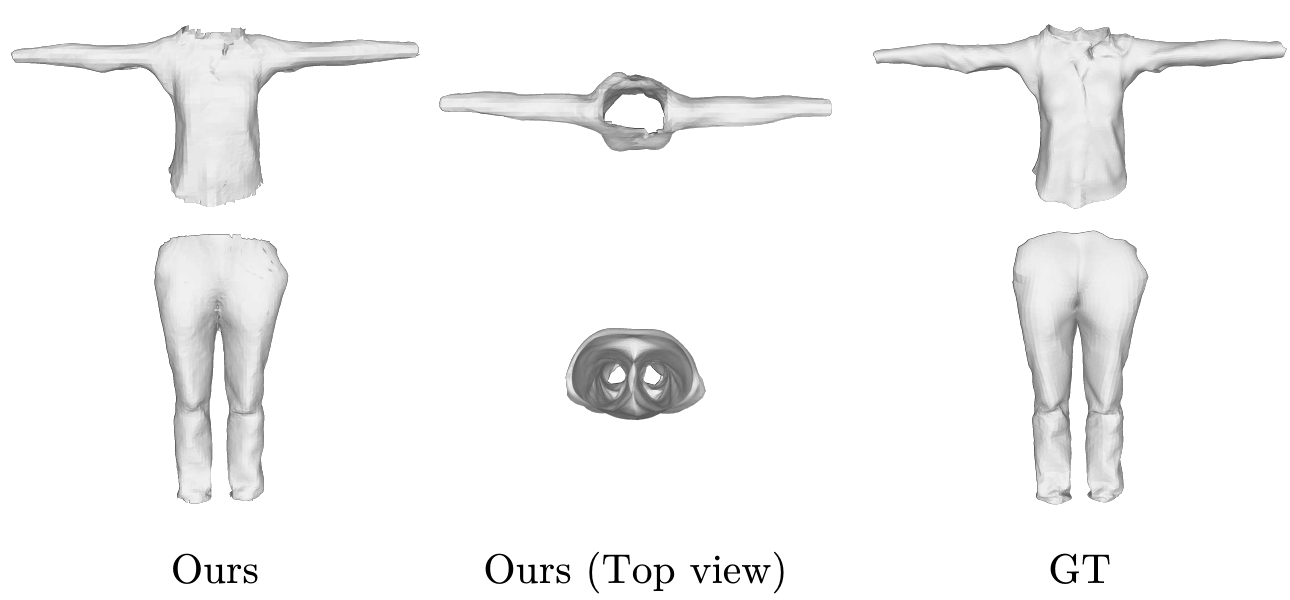}
  \vspace{-0.1in}
  \caption{\textbf{Reconstruction results on the MGN~\cite{bhatnagar2019multi} dataset.} Our method allows the reconstruction of the non-watertight garments.}
  \label{fig: general3}
\end{figure}

We also apply our model to garment reconstruction. Without retraining nor fine-tuning the model, we directly inference on MGN~\cite{bhatnagar2019multi} dataset. The reconstruction results are shown in Figure~\ref{fig: general3}. From the top view, we can see the garments are non-watertight as well.

\begin{table}[t]
  \centering
  \begin{tabular}{l|cc|cc}
    & \multicolumn{2}{c|}{Chamfer distance $\downarrow$} & \multicolumn{2}{c}{F-Score $\uparrow$} \\
    Method & Mean & Median  & F1\textsuperscript{0.005} & F1\textsuperscript{0.01} \\
    \shline
    Input & 0.363 & 0.355 & 48.50 & 88.34 \\
    Watertight GT & 2.628 & 2.293 & 68.82 & 81.60 \\
    NDF~\cite{DBLP:conf/nips/ChibaneMP20} & 0.126 & 0.120 & 88.09 & 99.54 \\
    Ours & 0.128 & 0.123 & 88.05 & 99.31 \\
  \end{tabular}
  \caption{\textbf{Quantitative evaluation on general shapes.} We train and evaluate our method on the raw data of the ShapeNet ``Car" category. Our method achieves comparable performance with the state-of-the-art method.}
  \label{tab: general}
\end{table}

\noindent
\textbf{Quantitative evaluation.} As shown in Table~\ref{tab: general}, we also quantitatively evaluate our method and find that GIFS achieves comparable performance with the state-of-the-art method NDF. Here, the evaluation of NDF is performed on point clouds without topologies rather than meshes. We also sample points from the watertight ground truth shapes and report its metric (watertight GT in table). We consider this as the upper bound of the traditional implicit function.

\noindent
\textbf{Speed comparison.} The limitation of NDF is not only the lack of continuity on the surface but also the extremely slow running speed. This is due to that the ball-pivoting~\cite{bernardini1999ball} algorithm used by NDF requires a large number of nearest-neighbor searches, normal calculation, and surface fitting. In the experiment, we use the default $1 \times 10^6$ points and the parameters provided by NDF.\footnote{Using downsampled points or different parameters will reduce the runtime, but it is also more likely to create non-smooth surfaces and holes.} The platform for NDF mesh conversion is a PC with a 6-Core 2.6 GHz CPU and the platform for the network inference is the RTX 2080 Ti. We report the runtime of NDF and our method in Table~\ref{tab: speed}. Note that the runtime here consists of both network inference and mesh generation. With the unoptimized code, our method has a tremendous advantage over the NDF.

\noindent
\textbf{Space comparison.} Similarly, a large number of points or the complex mesh generated by the ball-pivoting algorithm is also spatially inefficient. We statistics the average size of the files generated by both methods and report the result in Table~\ref{tab: speed}.

\subsection{Ablation Study}
\label{subsec: ablation}

\begin{table}[t]
  \centering
  \begin{tabular}{l|c|c}
    Method & Runtime $\downarrow$ & Space $\downarrow$ \\
    \shline
    NDF~\cite{DBLP:conf/nips/ChibaneMP20} & 2593 s & 40.57 MB \\
    Ours & \textbf{53 s} & \textbf{2.43 MB} \\
  \end{tabular}
  \caption{\textbf{Algorithm efficiency analysis.} Our method has a tremendous improvement in temporal and spatial efficiency over the previous method without losing accuracy. }
  \label{tab: speed}

\end{table}

\noindent
\textbf{Architecture.} We mainly ablate the UDF branch and the mesh refinement and report the results in Table~\ref{tab: ablation}. The experiment is conducted on watertight ShapeNet data, and the resolution of the encoding voxel grid is $128^3$. We observe that both modules play an important role in our framework. The unary UDF loss is complementary to our pair-wise  GIFS loss. The mesh refinement enables the generated mesh to be closer to the decision boundary.

\noindent
\textbf{Surface extraction.} We evaluate the performance of different 3D grid sizes in surface extraction. Table~\ref{tab: scalability} shows how the accuracy and speed change with the different resolutions of the 3D grid. We achieve better reconstruction with a higher resolution of the 3D grid. Thanks to the coarse-to-fine paradigm, the memory and computation cost do not have an exponential growth but instead only increase for a reasonable amount.

\noindent
\textbf{Sampling strategy.} We compare the performance of different $\sigma$ during data generation (see Sec~\ref{subsec: setting}) and report the results in Table~\ref{tab: sampling}. Our method is robust to different sampling strategies.

\begin{table}[t]
  \centering
  \begin{tabular}{l|cc|cc}
    & \multicolumn{2}{c|}{Chamfer distance $\downarrow$} & \multicolumn{2}{c}{F-Score $\uparrow$} \\
    Method & Mean  & Median  & F1\textsuperscript{0.005} & F1\textsuperscript{0.01} \\
    \shline
    w/o UDF & 0.227 & 0.206 & 73.38 & 97.42 \\
    w/o Refine & 0.152 & 0.121 & 87.98 & 97.93 \\
    Full model & \textbf{0.146} & \textbf{0.114} & \textbf{88.75} & \textbf{98.09}
  \end{tabular}
  \caption{\textbf{Ablation study.} Both UDF branch and mesh refinement boost the performance. }
  \label{tab: ablation}

\end{table}

\begin{table}[t]
  \vspace{-10pt}
  \setlength\tabcolsep{4pt}
  \centering
  \begin{tabular}{l|cc|cc|cc}
    & \multicolumn{2}{c|}{CD $\downarrow$} & \multicolumn{2}{c|}{F-score $\uparrow$} & \multicolumn{2}{c}{Efficiency $\downarrow$} \\
    Res & Mean  & Median  & F1\textsuperscript{0.005} & F1\textsuperscript{0.01} & Time & Space\\
    \shline
    $80^3$ & 0.445 & 0.231 & 70.30 & 97.66 & 12 {\footnotesize S} & 0.43 {\footnotesize MB} \\
    $160^3$ & 0.128 & 0.123 & 88.05 & 99.31 & 53 {\footnotesize S} & 2.43 {\footnotesize MB} \\
    $240^3$ & 0.125 & 0.121 &88.83 &99.36 &121 {\footnotesize S} &7.00 {\footnotesize MB} \\
    $320^3$ &0.123 & 0.119 &89.10 &99.36 &182 {\footnotesize S} &11.51 {\footnotesize MB} \\
  \end{tabular}
  \caption{\textbf{Surface extraction.} We compare the performance of different 3D grid sizes in surface extraction. Our method achieves better reconstruction with a higher resolution of the 3D grid.}
  \label{tab: scalability}
  \vspace{-10pt}
\end{table}

\begin{table}[t]
  \setlength\tabcolsep{5pt}
  \centering
  \begin{tabular}{l|cc|cc}
    & \multicolumn{2}{c|}{Chamfer distance $\downarrow$} & \multicolumn{2}{c}{F-score $\uparrow$}  \\
    $\sigma$  & Mean  & Median  & F1\textsuperscript{0.005} & F1\textsuperscript{0.01}\\
    \shline
    { 0.005, 0.01, 0.03} & 0.128 & 0.123 & 88.05 & 99.31 \\
    { 0.003, 0.01, 0.03} & 0.131 & 0.126 & 87.22 & 99.37 \\
    { 0.01, 0.03 } & 0.137 & 0.131 & 87.36 & 98.90 \\
  \end{tabular}
  \caption{\textbf{Sampling strategy.}  We compare the performance of different $\sigma$ during data generation (see Sec~\ref{subsec: setting}). Our method is robust to different sampling strategies.}
  \label{tab: sampling}
  \vspace{-0.1in}
\end{table}

\section{Discussion}

\noindent
\textbf{Conclusion.} In this paper, we introduce a generalized shape representation and its corresponding neural network GIFS. GIFS allows the reconstruction of high-quality general object shapes including watertight, non-watertight, and multi-layer shapes. We also devise an algorithm to directly extract mesh from GIFS. Experiments demonstrate that GIFS not only achieves state-of-the-art performance in both watertight and general shapes but also shows advantages in terms of efficiency and visual effect. We believe that GIFS takes a step towards general shape representation.

\noindent
\textbf{Limitations and future work.} We describe two limitations: (i) In our surface extraction algorithm, an assumption exists that at sufficiently small scales the surface can be divided into two classes. For general shapes, however, the assumption does not guarantee to hold when the resolution of the grid is low. (ii) Our running speed, limited by surface extraction based on brute-force strategy and simple loops, is still far from real-time. We also point out two possible research directions: (i) Improving the smoothness of the generated surface using the first and second-order gradients of the implicit field. (ii) Accelerating network inference and surface extraction using parallel computing.

\noindent
{\textbf{Acknowledgements.} This work was supported, in part, by gifts from TuSimple.}

\newpage

{\small
  \bibliographystyle{ieee_fullname}
  \bibliography{egbib}

\begin{thebibliography}{10}\itemsep=-1pt

\bibitem{alldieck2019tex2shape}
Thiemo Alldieck, Gerard Pons-Moll, Christian Theobalt, and Marcus Magnor.
\newblock Tex2shape: Detailed full human body geometry from a single image.
\newblock In {\em Proceedings of the IEEE/CVF International Conference on
  Computer Vision}, pages 2293--2303, 2019.

\bibitem{cgal:atw-aabb-21b}
Pierre Alliez, St{\'e}phane Tayeb, and Camille Wormser.
\newblock {3D} fast intersection and distance computation.
\newblock In {\em {CGAL} User and Reference Manual}. {CGAL Editorial Board},
  {5.3} edition, 2021.

\bibitem{atzmon2020sal}
Matan Atzmon and Yaron Lipman.
\newblock Sal: Sign agnostic learning of shapes from raw data.
\newblock In {\em Proceedings of the IEEE/CVF Conference on Computer Vision and
  Pattern Recognition}, pages 2565--2574, 2020.

\bibitem{atzmon2020sald}
Matan Atzmon and Yaron Lipman.
\newblock Sald: Sign agnostic learning with derivatives.
\newblock {\em arXiv preprint arXiv:2006.05400}, 2020.

\bibitem{bernardini1999ball}
Fausto Bernardini, Joshua Mittleman, Holly Rushmeier, Claudio Silva, and
  Gabriel Taubin.
\newblock The ball-pivoting algorithm for surface reconstruction.
\newblock {\em IEEE transactions on visualization and computer graphics},
  5(4):349--359, 1999.

\bibitem{bhatnagar2019multi}
Bharat~Lal Bhatnagar, Garvita Tiwari, Christian Theobalt, and Gerard Pons-Moll.
\newblock Multi-garment net: Learning to dress 3d people from images.
\newblock In {\em Proceedings of the IEEE/CVF international conference on
  computer vision}, pages 5420--5430, 2019.

\bibitem{bronstein2017geometric}
Michael~M Bronstein, Joan Bruna, Yann LeCun, Arthur Szlam, and Pierre
  Vandergheynst.
\newblock Geometric deep learning: going beyond euclidean data.
\newblock {\em IEEE Signal Processing Magazine}, 34(4):18--42, 2017.

\bibitem{calakli2011ssd}
Fatih Calakli and Gabriel Taubin.
\newblock Ssd: Smooth signed distance surface reconstruction.
\newblock In {\em Computer Graphics Forum}, volume~30, pages 1993--2002. Wiley
  Online Library, 2011.

\bibitem{chabra2020deep}
Rohan Chabra, Jan~E Lenssen, Eddy Ilg, Tanner Schmidt, Julian Straub, Steven
  Lovegrove, and Richard Newcombe.
\newblock Deep local shapes: Learning local sdf priors for detailed 3d
  reconstruction.
\newblock In {\em European Conference on Computer Vision}, pages 608--625,
  2020.

\bibitem{chang2015shapenet}
Angel~X Chang, Thomas Funkhouser, Leonidas Guibas, Pat Hanrahan, Qixing Huang,
  Zimo Li, Silvio Savarese, Manolis Savva, Shuran Song, Hao Su, et~al.
\newblock Shapenet: An information-rich 3d model repository.
\newblock {\em arXiv preprint arXiv:1512.03012}, 2015.

\bibitem{chen2019learning}
Zhiqin Chen and Hao Zhang.
\newblock Learning implicit fields for generative shape modeling.
\newblock In {\em Proceedings of the IEEE/CVF Conference on Computer Vision and
  Pattern Recognition}, pages 5939--5948, 2019.

\bibitem{chibane2020implicit}
Julian Chibane, Thiemo Alldieck, and Gerard Pons-Moll.
\newblock Implicit functions in feature space for 3d shape reconstruction and
  completion.
\newblock In {\em Proceedings of the IEEE/CVF Conference on Computer Vision and
  Pattern Recognition}, pages 6970--6981, 2020.

\bibitem{DBLP:conf/nips/ChibaneMP20}
Julian Chibane, Aymen Mir, and Gerard Pons{-}Moll.
\newblock Neural unsigned distance fields for implicit function learning.
\newblock In {\em Advances in Neural Information Processing Systems}, 2020.

\bibitem{choy20163d}
Christopher~B Choy, Danfei Xu, JunYoung Gwak, Kevin Chen, and Silvio Savarese.
\newblock 3d-r2n2: A unified approach for single and multi-view 3d object
  reconstruction.
\newblock In {\em European conference on computer vision}, pages 628--644,
  2016.

\bibitem{curless1996volumetric}
Brian Curless and Marc Levoy.
\newblock A volumetric method for building complex models from range images.
\newblock In {\em Proceedings of the 23rd annual conference on Computer
  graphics and interactive techniques}, pages 303--312, 1996.

\bibitem{dai2019scan2mesh}
Angela Dai and Matthias Nie{\ss}ner.
\newblock Scan2mesh: From unstructured range scans to 3d meshes.
\newblock In {\em Proceedings of the IEEE/CVF Conference on Computer Vision and
  Pattern Recognition}, pages 5574--5583, 2019.

\bibitem{dai2017shape}
Angela Dai, Charles Ruizhongtai~Qi, and Matthias Nie{\ss}ner.
\newblock Shape completion using 3d-encoder-predictor cnns and shape synthesis.
\newblock In {\em Proceedings of the IEEE Conference on Computer Vision and
  Pattern Recognition}, pages 5868--5877, 2017.

\bibitem{duan2020curriculum}
Yueqi Duan, Haidong Zhu, He Wang, Li Yi, Ram Nevatia, and Leonidas~J Guibas.
\newblock Curriculum deepsdf.
\newblock In {\em European Conference on Computer Vision}, pages 51--67, 2020.

\bibitem{fan2017point}
Haoqiang Fan, Hao Su, and Leonidas~J Guibas.
\newblock A point set generation network for 3d object reconstruction from a
  single image.
\newblock In {\em Proceedings of the IEEE conference on computer vision and
  pattern recognition}, pages 605--613, 2017.

\bibitem{fan2020pstnet}
Hehe Fan, Xin Yu, Yuhang Ding, Yi Yang, and Mohan Kankanhalli.
\newblock Pstnet: Point spatio-temporal convolution on point cloud sequences.
\newblock In {\em International conference on learning representations}, 2020.

\bibitem{DBLP:conf/cvpr/GenovaCSSF20}
Kyle Genova, Forrester Cole, Avneesh Sud, Aaron Sarna, and Thomas~A.
  Funkhouser.
\newblock Local deep implicit functions for 3d shape.
\newblock In {\em CVPR}, pages 4856--4865, 2020.

\bibitem{genova2019learning}
Kyle Genova, Forrester Cole, Daniel Vlasic, Aaron Sarna, William~T Freeman, and
  Thomas Funkhouser.
\newblock Learning shape templates with structured implicit functions.
\newblock In {\em Proceedings of the IEEE/CVF International Conference on
  Computer Vision}, pages 7154--7164, 2019.

\bibitem{gkioxari2019mesh}
Georgia Gkioxari, Jitendra Malik, and Justin Johnson.
\newblock Mesh r-cnn.
\newblock In {\em Proceedings of the IEEE/CVF International Conference on
  Computer Vision}, pages 9785--9795, 2019.

\bibitem{gropp2020implicit}
Amos Gropp, Lior Yariv, Niv Haim, Matan Atzmon, and Yaron Lipman.
\newblock Implicit geometric regularization for learning shapes.
\newblock {\em arXiv preprint arXiv:2002.10099}, 2020.

\bibitem{groueix2018papier}
Thibault Groueix, Matthew Fisher, Vladimir~G Kim, Bryan~C Russell, and Mathieu
  Aubry.
\newblock A papier-m{\^a}ch{\'e} approach to learning 3d surface generation.
\newblock In {\em Proceedings of the IEEE conference on computer vision and
  pattern recognition}, pages 216--224, 2018.

\bibitem{guo20153d}
Kan Guo, Dongqing Zou, and Xiaowu Chen.
\newblock 3d mesh labeling via deep convolutional neural networks.
\newblock {\em ACM Transactions on Graphics (TOG)}, 35(1):1--12, 2015.

\bibitem{guo2021pct}
Meng-Hao Guo, Jun-Xiong Cai, Zheng-Ning Liu, Tai-Jiang Mu, Ralph~R Martin, and
  Shi-Min Hu.
\newblock Pct: Point cloud transformer.
\newblock {\em Computational Visual Media}, 7(2):187--199, 2021.

\bibitem{hane2017hierarchical}
Christian H{\"a}ne, Shubham Tulsiani, and Jitendra Malik.
\newblock Hierarchical surface prediction for 3d object reconstruction.
\newblock In {\em 2017 International Conference on 3D Vision (3DV)}, pages
  412--420. IEEE, 2017.

\bibitem{Hui2021PyramidPC}
Le Hui, Hang Yang, Mingmei Cheng, Jin Xie, and Jian Yang.
\newblock Pyramid point cloud transformer for large-scale place recognition.
\newblock {\em 2021 IEEE/CVF International Conference on Computer Vision
  (ICCV)}, pages 6078--6087, 2021.

\bibitem{ji2017surfacenet}
Mengqi Ji, Juergen Gall, Haitian Zheng, Yebin Liu, and Lu Fang.
\newblock Surfacenet: An end-to-end 3d neural network for multiview stereopsis.
\newblock In {\em Proceedings of the IEEE International Conference on Computer
  Vision}, pages 2307--2315, 2017.

\bibitem{jiang2020local}
Chiyu Jiang, Avneesh Sud, Ameesh Makadia, Jingwei Huang, Matthias Nie{\ss}ner,
  Thomas Funkhouser, et~al.
\newblock Local implicit grid representations for 3d scenes.
\newblock In {\em Proceedings of the IEEE/CVF Conference on Computer Vision and
  Pattern Recognition}, pages 6001--6010, 2020.

\bibitem{jimenez2016unsupervised}
Danilo Jimenez~Rezende, SM Eslami, Shakir Mohamed, Peter Battaglia, Max
  Jaderberg, and Nicolas Heess.
\newblock Unsupervised learning of 3d structure from images.
\newblock {\em Advances in Neural Information Processing Systems},
  29:4996--5004, 2016.

\bibitem{kanazawa2018end}
Angjoo Kanazawa, Michael~J Black, David~W Jacobs, and Jitendra Malik.
\newblock End-to-end recovery of human shape and pose.
\newblock In {\em Proceedings of the IEEE conference on computer vision and
  pattern recognition}, pages 7122--7131, 2018.

\bibitem{kanazawa2018learning}
Angjoo Kanazawa, Shubham Tulsiani, Alexei~A Efros, and Jitendra Malik.
\newblock Learning category-specific mesh reconstruction from image
  collections.
\newblock In {\em Proceedings of the European Conference on Computer Vision
  (ECCV)}, pages 371--386, 2018.

\bibitem{DBLP:conf/nips/KarHM17}
Abhishek Kar, Christian H{\"{a}}ne, and Jitendra Malik.
\newblock Learning a multi-view stereo machine.
\newblock In {\em Advances in Neural Information Processing Systems}, pages
  365--376, 2017.

\bibitem{kazhdan2006poisson}
Michael Kazhdan, Matthew Bolitho, and Hugues Hoppe.
\newblock Poisson surface reconstruction.
\newblock In {\em Proceedings of the fourth Eurographics symposium on Geometry
  processing}, volume~7, 2006.

\bibitem{kazhdan2013screened}
Michael Kazhdan and Hugues Hoppe.
\newblock Screened poisson surface reconstruction.
\newblock {\em ACM Transactions on Graphics (ToG)}, 32(3):1--13, 2013.

\bibitem{kleineberg2020adversarial}
Marian Kleineberg, Matthias Fey, and Frank Weichert.
\newblock Adversarial generation of continuous implicit shape representations.
\newblock {\em arXiv preprint arXiv:2002.00349}, 2020.

\bibitem{kolotouros2019learning}
Nikos Kolotouros, Georgios Pavlakos, Michael~J Black, and Kostas Daniilidis.
\newblock Learning to reconstruct 3d human pose and shape via model-fitting in
  the loop.
\newblock In {\em Proceedings of the IEEE/CVF International Conference on
  Computer Vision}, pages 2252--2261, 2019.

\bibitem{kolotouros2019convolutional}
Nikos Kolotouros, Georgios Pavlakos, and Kostas Daniilidis.
\newblock Convolutional mesh regression for single-image human shape
  reconstruction.
\newblock In {\em Proceedings of the IEEE/CVF Conference on Computer Vision and
  Pattern Recognition}, pages 4501--4510, 2019.

\bibitem{ladicky2017point}
Lubor Ladicky, Olivier Saurer, SoHyeon Jeong, Fabio Maninchedda, and Marc
  Pollefeys.
\newblock From point clouds to mesh using regression.
\newblock In {\em Proceedings of the IEEE International Conference on Computer
  Vision}, pages 3893--3902, 2017.

\bibitem{landrieu2018large}
Loic Landrieu and Martin Simonovsky.
\newblock Large-scale point cloud semantic segmentation with superpoint graphs.
\newblock In {\em Proceedings of the IEEE conference on computer vision and
  pattern recognition}, pages 4558--4567, 2018.

\bibitem{liao2018deep}
Yiyi Liao, Simon Donne, and Andreas Geiger.
\newblock Deep marching cubes: Learning explicit surface representations.
\newblock In {\em Proceedings of the IEEE Conference on Computer Vision and
  Pattern Recognition}, pages 2916--2925, 2018.

\bibitem{lin2019photometric}
Chen-Hsuan Lin, Oliver Wang, Bryan~C Russell, Eli Shechtman, Vladimir~G Kim,
  Matthew Fisher, and Simon Lucey.
\newblock Photometric mesh optimization for video-aligned 3d object
  reconstruction.
\newblock In {\em Proceedings of the IEEE/CVF Conference on Computer Vision and
  Pattern Recognition}, pages 969--978, 2019.

\bibitem{loper2015smpl}
Matthew Loper, Naureen Mahmood, Javier Romero, Gerard Pons-Moll, and Michael~J
  Black.
\newblock Smpl: A skinned multi-person linear model.
\newblock {\em ACM transactions on graphics (TOG)}, 34(6):1--16, 2015.

\bibitem{lorensen1987marching}
William~E Lorensen and Harvey~E Cline.
\newblock Marching cubes: A high resolution 3d surface construction algorithm.
\newblock {\em ACM siggraph computer graphics}, 21(4):163--169, 1987.

\bibitem{mescheder2019occupancy}
Lars Mescheder, Michael Oechsle, Michael Niemeyer, Sebastian Nowozin, and
  Andreas Geiger.
\newblock Occupancy networks: Learning 3d reconstruction in function space.
\newblock In {\em Proceedings of the IEEE/CVF Conference on Computer Vision and
  Pattern Recognition}, pages 4460--4470, 2019.

\bibitem{park2019deepsdf}
Jeong~Joon Park, Peter Florence, Julian Straub, Richard Newcombe, and Steven
  Lovegrove.
\newblock Deepsdf: Learning continuous signed distance functions for shape
  representation.
\newblock In {\em Proceedings of the IEEE/CVF Conference on Computer Vision and
  Pattern Recognition}, pages 165--174, 2019.

\bibitem{pons2017clothcap}
Gerard Pons-Moll, Sergi Pujades, Sonny Hu, and Michael~J Black.
\newblock Clothcap: Seamless 4d clothing capture and retargeting.
\newblock {\em ACM Transactions on Graphics (ToG)}, 36(4):1--15, 2017.

\bibitem{qi2017pointnet}
Charles~R Qi, Hao Su, Kaichun Mo, and Leonidas~J Guibas.
\newblock Pointnet: Deep learning on point sets for 3d classification and
  segmentation.
\newblock In {\em Proceedings of the IEEE conference on computer vision and
  pattern recognition}, pages 652--660, 2017.

\bibitem{DBLP:conf/nips/QiYSG17}
Charles~Ruizhongtai Qi, Li Yi, Hao Su, and Leonidas~J. Guibas.
\newblock Pointnet++: Deep hierarchical feature learning on point sets in a
  metric space.
\newblock In {\em Advances in Neural Information Processing Systems}, pages
  5099--5108, 2017.

\bibitem{ranjan2018generating}
Anurag Ranjan, Timo Bolkart, Soubhik Sanyal, and Michael~J Black.
\newblock Generating 3d faces using convolutional mesh autoencoders.
\newblock In {\em Proceedings of the European Conference on Computer Vision
  (ECCV)}, pages 704--720, 2018.

\bibitem{saito2019pifu}
Shunsuke Saito, Zeng Huang, Ryota Natsume, Shigeo Morishima, Angjoo Kanazawa,
  and Hao Li.
\newblock Pifu: Pixel-aligned implicit function for high-resolution clothed
  human digitization.
\newblock In {\em Proceedings of the IEEE/CVF International Conference on
  Computer Vision}, pages 2304--2314, 2019.

\bibitem{saito2020pifuhd}
Shunsuke Saito, Tomas Simon, Jason Saragih, and Hanbyul Joo.
\newblock Pifuhd: Multi-level pixel-aligned implicit function for
  high-resolution 3d human digitization.
\newblock In {\em Proceedings of the IEEE/CVF Conference on Computer Vision and
  Pattern Recognition}, pages 84--93, 2020.

\bibitem{sitzmann2020implicit}
Vincent Sitzmann, Julien Martel, Alexander Bergman, David Lindell, and Gordon
  Wetzstein.
\newblock Implicit neural representations with periodic activation functions.
\newblock {\em Advances in Neural Information Processing Systems}, 33, 2020.

\bibitem{DBLP:conf/nips/TancikSMFRSRBN20}
Matthew Tancik, Pratul~P. Srinivasan, Ben Mildenhall, Sara Fridovich{-}Keil,
  Nithin Raghavan, Utkarsh Singhal, Ravi Ramamoorthi, Jonathan~T. Barron, and
  Ren Ng.
\newblock Fourier features let networks learn high frequency functions in low
  dimensional domains.
\newblock In {\em Advances in Neural Information Processing Systems}, 2020.

\bibitem{tatarchenko2017octree}
Maxim Tatarchenko, Alexey Dosovitskiy, and Thomas Brox.
\newblock Octree generating networks: Efficient convolutional architectures for
  high-resolution 3d outputs.
\newblock In {\em Proceedings of the IEEE International Conference on Computer
  Vision}, pages 2088--2096, 2017.

\bibitem{thomas2019kpconv}
Hugues Thomas, Charles~R Qi, Jean-Emmanuel Deschaud, Beatriz Marcotegui,
  Fran{\c{c}}ois Goulette, and Leonidas~J Guibas.
\newblock Kpconv: Flexible and deformable convolution for point clouds.
\newblock In {\em Proceedings of the IEEE/CVF International Conference on
  Computer Vision}, pages 6411--6420, 2019.

\bibitem{venkatesh2021deep}
Rahul Venkatesh, Tejan Karmali, Sarthak Sharma, Aurobrata Ghosh, R~Venkatesh
  Babu, L{\'a}szl{\'o}~A Jeni, and Maneesh Singh.
\newblock Deep implicit surface point prediction networks.
\newblock In {\em Proceedings of the IEEE/CVF International Conference on
  Computer Vision}, pages 12653--12662, 2021.

\bibitem{wang2018pixel2mesh}
Nanyang Wang, Yinda Zhang, Zhuwen Li, Yanwei Fu, Wei Liu, and Yu-Gang Jiang.
\newblock Pixel2mesh: Generating 3d mesh models from single rgb images.
\newblock In {\em Proceedings of the European Conference on Computer Vision
  (ECCV)}, pages 52--67, 2018.

\bibitem{wang2019dynamic}
Yue Wang, Yongbin Sun, Ziwei Liu, Sanjay~E Sarma, Michael~M Bronstein, and
  Justin~M Solomon.
\newblock Dynamic graph cnn for learning on point clouds.
\newblock {\em Acm Transactions On Graphics (tog)}, 38(5):1--12, 2019.

\bibitem{DBLP:conf/nips/0001WXSFT17}
Jiajun Wu, Yifan Wang, Tianfan Xue, Xingyuan Sun, Bill Freeman, and Josh
  Tenenbaum.
\newblock Marrnet: 3d shape reconstruction via 2.5d sketches.
\newblock In {\em Advances in Neural Information Processing Systems}, pages
  540--550, 2017.

\bibitem{wu2016learning}
Jiajun Wu, Chengkai Zhang, Tianfan Xue, William~T Freeman, and Joshua~B
  Tenenbaum.
\newblock Learning a probabilistic latent space of object shapes via 3d
  generative-adversarial modeling.
\newblock In {\em Advances in Neural Information Processing Systems}, pages
  82--90, 2016.

\bibitem{wu2019pointconv}
Wenxuan Wu, Zhongang Qi, and Li Fuxin.
\newblock Pointconv: Deep convolutional networks on 3d point clouds.
\newblock In {\em Proceedings of the IEEE/CVF Conference on Computer Vision and
  Pattern Recognition}, pages 9621--9630, 2019.

\bibitem{wu20153d}
Zhirong Wu, Shuran Song, Aditya Khosla, Fisher Yu, Linguang Zhang, Xiaoou Tang,
  and Jianxiong Xiao.
\newblock 3d shapenets: A deep representation for volumetric shapes.
\newblock In {\em Proceedings of the IEEE conference on computer vision and
  pattern recognition}, pages 1912--1920, 2015.

\bibitem{xu2019disn}
Qiangeng Xu, Weiyue Wang, Duygu Ceylan, Radomir Mech, and Ulrich Neumann.
\newblock Disn: Deep implicit surface network for high-quality single-view 3d
  reconstruction.
\newblock {\em arXiv preprint arXiv:1905.10711}, 2019.

\bibitem{zhao2021learning}
Fang Zhao, Wenhao Wang, Shengcai Liao, and Ling Shao.
\newblock Learning anchored unsigned distance functions with gradient direction
  alignment for single-view garment reconstruction.
\newblock In {\em Proceedings of the IEEE/CVF International Conference on
  Computer Vision}, pages 12674--12683, 2021.

\bibitem{zhao2021sign}
Wenbin Zhao, Jiabao Lei, Yuxin Wen, Jianguo Zhang, and Kui Jia.
\newblock Sign-agnostic implicit learning of surface self-similarities for
  shape modeling and reconstruction from raw point clouds.
\newblock In {\em Proceedings of the IEEE/CVF Conference on Computer Vision and
  Pattern Recognition}, pages 10256--10265, 2021.

\end{thebibliography}
}

\newpage
\appendix

\section{Implementation Details}

\noindent
\textbf{Padding Issue.} In the original IF-Net implementation, the size of the 3D encoding grid is exactly the same as the normalized mesh. In experiments, we find that the lack of padding is prone to generate artifacts on the boundaries, which significantly degrades the reconstruction accuracy. In our implementation, the size of the normalized mesh is 0.9 of the encoding grid. Moreover, in 3D convolution, we find that the zero padding  outperforms the border padding used in IF-Net.

\noindent
\textbf{Training procedure.} During training, the number of training pairs is 50000 per instance and the batch size is 8. We employ the Adam optimizer with a learning rate of $1 \times 10^{-4}$ The watertight and the general shape experiments take 200 and 300 epochs respectively.

\noindent
\textbf{Mesh refinement.} The initial mesh produced by our adapted Marching Cubes is further refined by minimizing the UDF values on the mesh surface. We employ an RMSprop optimizer with an initial learning rate of $2 \times 10^{-4}$. In each iteration, a random point is sampled on each face of the mesh. Given a trained GIFS model, we take the sampled points as input, query and minimize their UDF values. The total number of iterations is 30.

\section{Surface Extraction Algorithm}

In this section, we provide the detailed surface extraction algorithm flow. Our algorithm consists of three steps: (i) Locate cubes that intersect the surface in a coarse-to-fine paradigm; (ii) Generate mesh triangles in final intersecting cubes with our adapted Marching Cubes; (iii) Refine mesh with the UDF branch.

First, we introduce our coarse-to-fine intersecting cubes localization algorithm. In our implementation, the initial resolution of the grid is $20^3$ and is subdivided 3 times. The final resolution is $160^3$. We show the detailed process in Algorithm~\ref{algo: locate}. Among the inputs of the algorithm, the initial intersecting indices $\boldsymbol{I}$ are integer indices of all $20^3$ cubes, the initial cube size $s_0 = 1.0 / 20 = 0.05$, the total number of stages $T = 3$ and the intersecting threshold $\tau = 2$.

After obtaining intersecting indices $\boldsymbol{I}$, the next step is to generate triangles using our adapted Marching Cubes. In each cube, we first use our model to predict all binary flags between 8 vertices, then assign binary labels (0/1) to 8 vertices based on the binary flags, and finally generate triangles with the lookup table provided by the original Marching Cubes. We show the detailed process in Algorithm~\ref{algo: mesh}. Among the inputs of the algorithm, $\boldsymbol{A} = \{0, 1\}^8$ is the all possible binary assignments for 8 vertices.

\begin{algorithm}[t]
  \small
  \renewcommand{\algorithmicrequire}{\textbf{Input:}}
  \renewcommand{\algorithmicensure}{\textbf{Output:}}
  \caption{Locate intersecting cubes}
  \label{algo: locate}
  \begin{algorithmic}[1]
    \REQUIRE Initial intersecting indices $\boldsymbol{I} \in \mathbb{Z}_{0}^{+ N \times 3}$, initial cube size $s_0$, point embedding network $g_{\boldsymbol{\theta}_1}$, UDF network $h_{\boldsymbol{\theta}_3}$, intersecting threshold $\tau$, total number of stages $T$.
    \ENSURE Intersecting indices $\boldsymbol{I} \in \mathbb{Z}_{0}^{+ M \times 3}$
    \FOR{stage $t \in range(T)$}
    \STATE Cube size $s \xleftarrow{}
      s_0 / 2^{t} $
    \STATE New empty intersecting indices $\boldsymbol{I}_n \xleftarrow{} \{\}$
    \FOR{intersecting index $\boldsymbol{i} \in \boldsymbol{I}$}
    \STATE Center of the intersecting cube $\boldsymbol{p}_c \xleftarrow{} s \boldsymbol{i}$
    \STATE Predicted UDF of the center $\boldsymbol{u} \xleftarrow{} h_{\boldsymbol{\theta}_3}(g_{\boldsymbol{\theta}_1}(\boldsymbol{p}_c)) $
    \IF{$\boldsymbol{u} < s \tau$}
    \STATE Subdivide current cube and add new indices to $\boldsymbol{I}_n$
    \ENDIF
    \ENDFOR
    \STATE  $\boldsymbol{I} \xleftarrow{} \boldsymbol{I}_n$
    \ENDFOR
  \end{algorithmic}
\end{algorithm}

\begin{algorithm}[t]
  \small
  \renewcommand{\algorithmicrequire}{\textbf{Input:}}
  \renewcommand{\algorithmicensure}{\textbf{Output:}}
  \caption{Adapted Marching Cubes}
  \label{algo: mesh}
  \begin{algorithmic}[1]
    \REQUIRE Intersecting indices $\boldsymbol{I} \in \mathbb{Z}_{0}^{+ M \times 3}$, cube size $s$, point embedding network $g_{\boldsymbol{\theta}_1}$, decoder $f_{\boldsymbol{\theta}_2}$, all possible assignments $\boldsymbol{A}$
    \ENSURE Mesh $\boldsymbol{M} = (\boldsymbol{V}, \boldsymbol{F})$

    \STATE Empty mesh $\boldsymbol{M} \xleftarrow{} \{\}$
    \FOR{intersecting index $\boldsymbol{i} \in \boldsymbol{I}$}
    \STATE Calculate 8 vertices of the intersecting cube using $\boldsymbol{i}$ and $s$.
    \STATE Predict 28 binary flags between 8 vertices using Eq.3

    \STATE Minimal cost $l_{min} \xleftarrow{} +\infty$
    \FOR{possible assignment $\boldsymbol{a} \in \boldsymbol{A}$}
    \STATE Calculate cost $l$ for assignment $\boldsymbol{a}$ using Eq.8
    \IF{$l < l_{min}$}
    \STATE $l_{min} \xleftarrow{} l$, $\boldsymbol{a}_{min} \xleftarrow{} \boldsymbol{a}$
    \ENDIF
    \ENDFOR
    \STATE Query the vertices and faces in the lookup table according to assignment $\boldsymbol{a}_{min}$ and add to $\boldsymbol{M}$
    \ENDFOR
  \end{algorithmic}
\end{algorithm}

\begin{algorithm}[t]
  \small
  \renewcommand{\algorithmicrequire}{\textbf{Input:}}
  \renewcommand{\algorithmicensure}{\textbf{Output:}}
  \caption{Mesh refinement}
  \label{algo: refine}
  \begin{algorithmic}[1]
    \REQUIRE  Mesh $\boldsymbol{M} = (\boldsymbol{V}, \boldsymbol{F})$, point embedding network $g_{\boldsymbol{\theta}_1}$, UDF network $h_{\boldsymbol{\theta}_3}$, number of iteration $N$
    \ENSURE Refined mesh $\boldsymbol{M} = (\boldsymbol{V}, \boldsymbol{F})$

    \FOR{iteration $n \in range(N)$}
    \STATE Sample points $\boldsymbol{P}$ from each face, each sampled point is a linear combination of 3 mesh vertices
    \STATE Optimize mesh vertices $\boldsymbol{V}$ by minimizing Eq.9
    \ENDFOR
  \end{algorithmic}
\end{algorithm}

\begin{figure*}[t]
  \centering
  \includegraphics[width=0.9\textwidth]{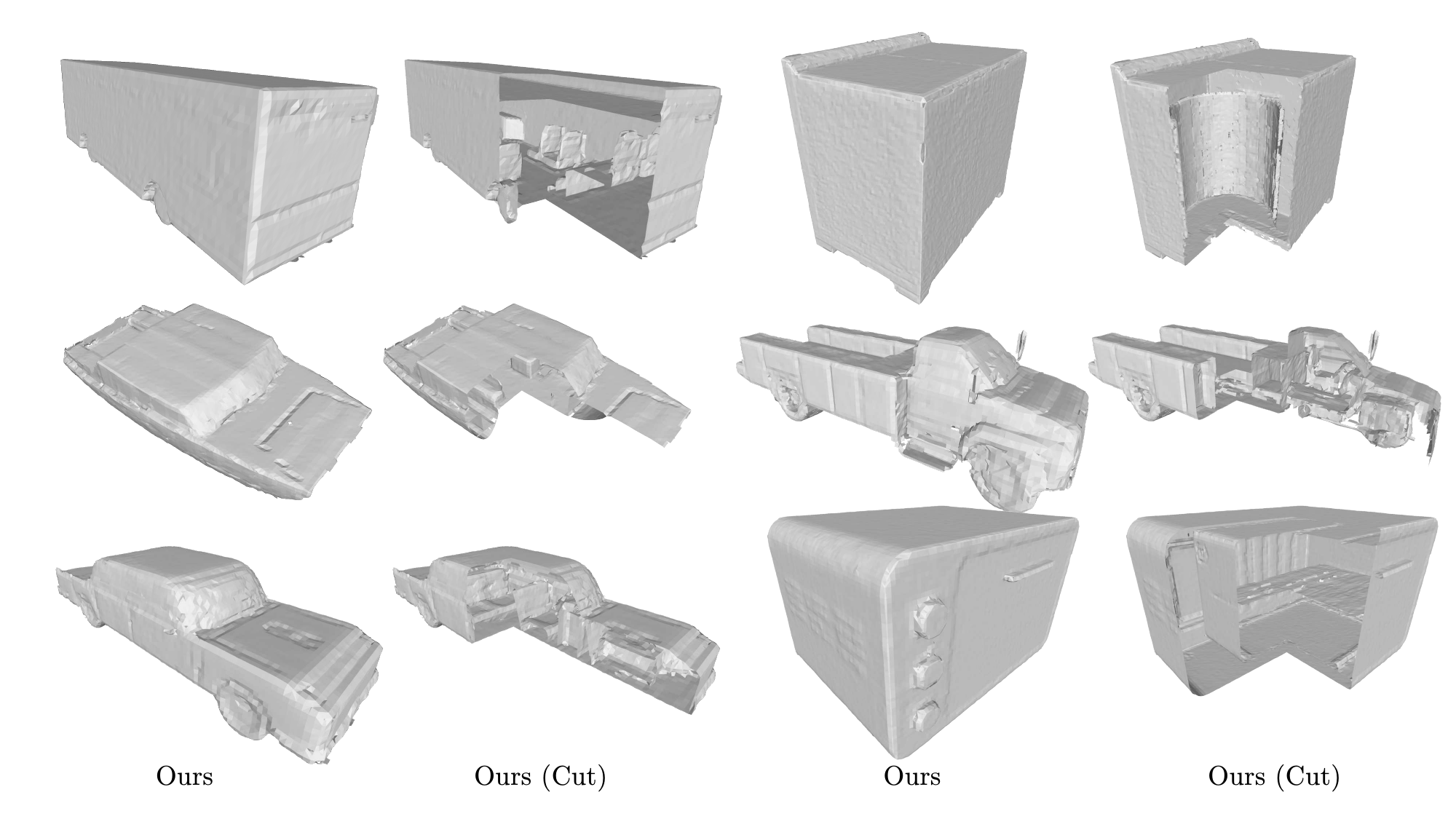}
  \caption{\textbf{Reconstruction results of multi-layer shapes.} Our method can reconstruct internal structures of various shapes. }
  \label{fig: supp0}
\end{figure*}

\begin{figure*}[t]
  \centering
  \includegraphics[width=0.9\textwidth]{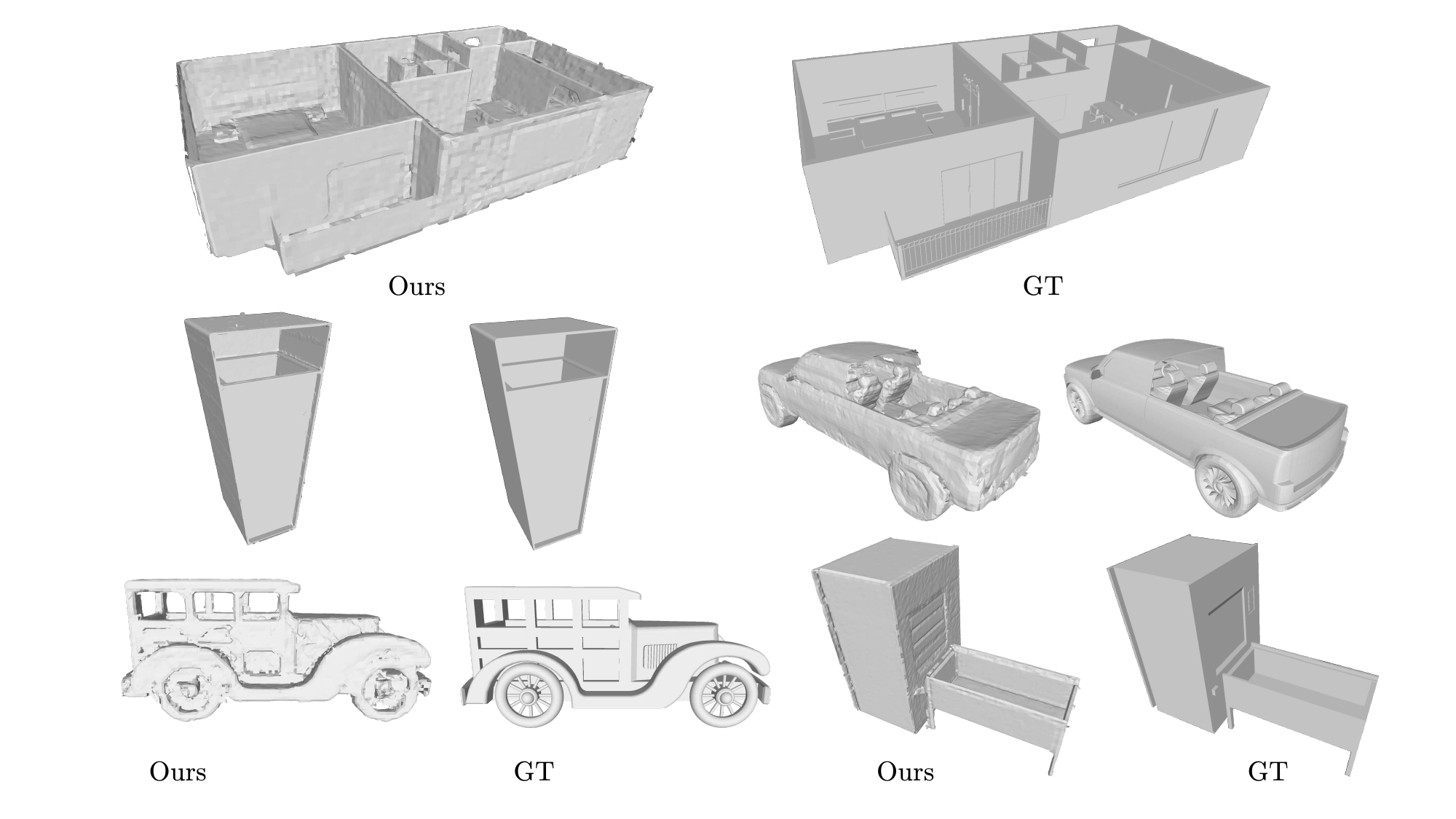}
  \caption{\textbf{Reconstruction results of non-watertight shapes.} The non-watertight shapes are difficult for traditional neural implicit functions to reconstruct. }
  \label{fig: supp1}
\end{figure*}

Finally, we utilize the UDF branch to refine the mesh $\boldsymbol{M} = (\boldsymbol{V}, \boldsymbol{F})$. Basically, we sample points on each face and refine mesh vertices by minimizing the UDF values of sampled points. We show the detailed process in Algorithm~\ref{algo: refine}. Among inputs, the number of iteration $N$ is 30.

\section{Qualitative Evaluation}

Reconstruction results of multi-layer shapes, non-watertight shapes and garments are shown in Figure~\ref{fig: supp0} and Figure~\ref{fig: supp1}.

\end{document}